%% file: acl_latex.tex
\documentclass[11pt]{article}
\PassOptionsToPackage{table}{xcolor}
\PassOptionsToPackage{disable}{todonotes}
\PassOptionsToPackage{bibliography=common}{apxproof}
\usepackage{todonotes}

\usepackage[preprint]{acl}

\usepackage{times}
\usepackage{latexsym}

\usepackage[T1]{fontenc}

\usepackage[utf8]{inputenc}

\usepackage{microtype}

\usepackage{inconsolata}

\usepackage{graphicx}
\usepackage{amsmath}
\usepackage{amssymb}
\usepackage{pifont}
\usepackage{xcolor}
\usepackage{empheq}
\usepackage{rycolab}
\usepackage{hexatagger}

\definecolor{tagcolor}{HTML}{B45F06}
\definecolor{relposcolor}{HTML}{2E7D32}
\definecolor{gatecolor}{HTML}{2E7D32} 

\usepackage{booktabs}
\usepackage{fontawesome5} 
\usepackage{tabularx}
\usepackage[table]{xcolor}
\usepackage{makecell}

\usepackage{arydshln} 

\usepackage{float}

\newcommand{\updelta}[1]{\textcolor{green!70!black}{$\uparrow$~+#1}}
\newcommand{\downdelta}[1]{\textcolor{red}{$\downarrow$~#1}}
\newcommand{\basedelta}{--}
\newcommand{\sectrow}[1]{%
\rowcolor{gray!15}
\multicolumn{7}{c}{\textit{#1}}\\
}
\newcommand{\res}[2]{\raisebox{-0.5\height}{\shortstack{#1\\[0.5pt]{\scriptsize(±#2)}}}}

\newcommand{\greencheck}{\textcolor{green}{\ding{51}}}
\newcommand{\redcross}{\textcolor{red}{\ding{55}}}

\definecolor{tagcolor}{HTML}{B45F06}
\definecolor{relposcolor}{HTML}{2E7D32}
\definecolor{sipecol}{RGB}{232,245,233} 
\definecolor{ppcol}{RGB}{232,240,254}   

\renewcommand{\todo}[1]{\textcolor{red}{TODO: #1}}

\title{Beyond Sequence Order: Syntax-Informed Positional Embeddings for Transformers}

\author{Haris Riaz \and Hyungji Kim \and Mihai Surdeanu \\
  Department of Computer Science \\
  University of Arizona \\
  Tucson, AZ, USA \\[0.5em]
  \faGithub\ \href{https://github.com/hriaz17/SiPE}{Code} \\}

\begin{document}
\maketitle
\begin{abstract}
Positional embeddings (PE) in Transformers encode token distance and order but are largely agnostic to \textit{syntactic structure}. We introduce \textbf{S}yntax-\textbf{i}nformed \textbf{P}ositional \textbf{E}mbeddings (\textbf{SiPE}), which learns a lightweight syntactic prior from dependency parses during pretraining and injects it across all three dominant PE families (absolute, relative, rotary), for both encoders and decoders, leaving self-attention and the rest of the architecture untouched. We isolate \emph{where} and \emph{how} the prior should enter the model, and find it depends on the architecture: for autoregressive decoders that use relative PE, the prior is strongest when coupled multiplicatively with the relative-position term of the attention score, outperforming injection into the input embeddings, into self-attention, or into the positional and attention terms jointly---while for encoders it is best added directly to the input embeddings, composing with each encoder's native positional mechanism. We find that models pre-trained with SiPE improve on the SyntaxGym benchmark by up to $10.3\%$ while simultaneously reducing perplexity by $9.0\%$ over a base model with no syntactic supervision---a metric nearly every existing syntax-injection method instead degrades. Crucially, these gains extend beyond syntactic generalization: SiPE also improves real-world language understanding, raising scores on the GLUE benchmark by up to $8.2\%$ over a model trained without it.
Unlike existing syntactic language models that marginalize over many parses at inference or discard syntax at runtime, SiPE conditions on a single parse, establishing a new Pareto frontier between syntactic supervision and inference cost.
\end{abstract}


\begin{figure}[t]
\centering
\includegraphics[width=\columnwidth]{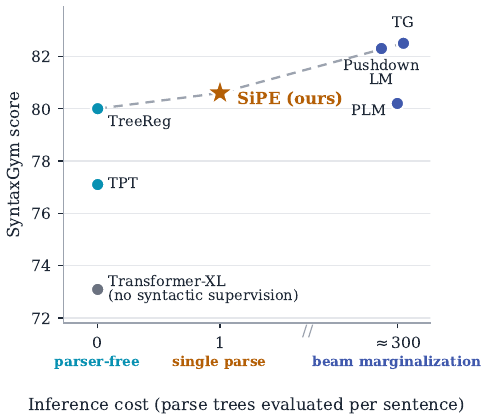}
\caption{\footnotesize\textbf{Our Syntax-Informed Positional Embeddings (SiPE) move the Pareto frontier between syntactic supervision and inference cost.} Each point plots a method's SyntaxGym score (\autoref{tab:results_delta-dup1}) against the number of parse trees it evaluates per sentence at inference. Joint syntactic LMs (TG~\cite{Sartran_2022}, Pushdown LM~\cite{murty2023pushdownlayersencodingrecursive}, PLM~\cite{qian-etal-2021-structural}) recover $p(x)$ by marginalizing over $\approx$300 candidate parses; parser-free methods (TreeReg~\cite{nandi2025sneakingsyntaxtransformerlanguage}, Tree-Planted Transformers~\cite{Yoshida_2024}) inject syntax only during training and discard the parser at runtime, making inference cheap but sacrificing some syntactic knowledge. SiPE conditions on a single parse, retaining most of the benefit of full marginalization at $1/300$th of its inference cost.}
\label{fig:pareto}
\end{figure}

\begin{figure*}[t]
  \centering
  \includegraphics[width=\textwidth]{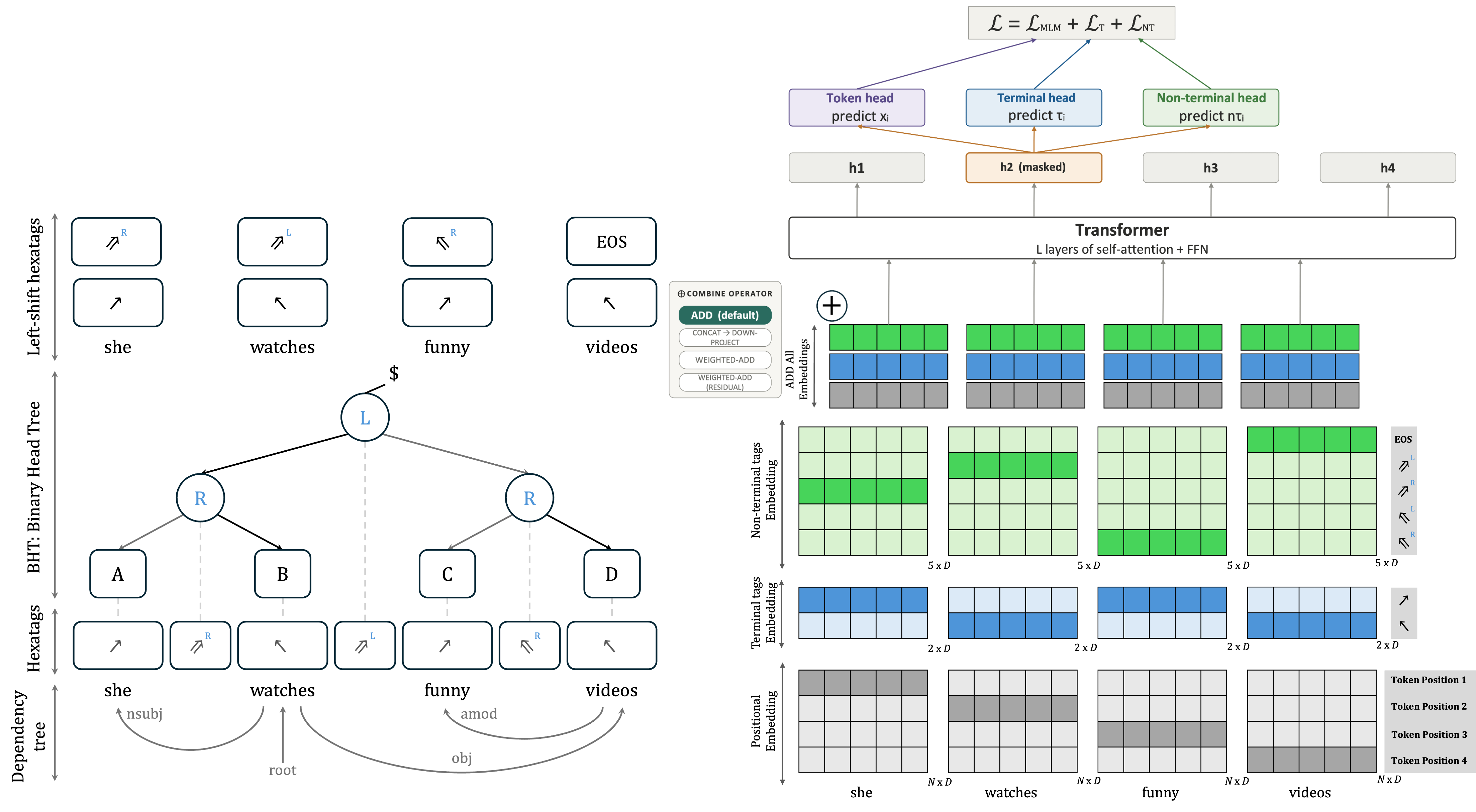}

\caption{ \footnotesize \textbf{Left:} From bottom to top, the figure shows the correspondence between dependency arcs, binary head tree (BHT), and hexatags (left-shifted) for the sentence ``she watches funny videos.'' \textbf{Right:} Our prior-injection method for absolute positional embeddings (input-pathway injection). From bottom to top, each token (at its first subword; \autoref{subword-tag-injection}) selects one row from each of four lookup tables: the token embedding ($|V| \times D$; omitted from the figure for brevity), the non-terminal tag embedding $\mathbf{E}^{N}_{\nu_p}$ ($5 \times D$), the terminal tag embedding $\mathbf{E}^{T}_{\tau_p}$ ($2 \times D$), and the positional embedding ($N \times D$). The selected rows are combined via an injection operation $\oplus$ (addition by default; we ablate concatenation, weighted-addition and addition to skip connection in \autoref{app:alternate-combination-appendix}) into the syntax-informed input $\mathbf{x}_p$ (\autoref{eq:roberta_input}). The transformer maps these inputs to contextual states $\mathbf{h}_i$, and at each masked position $i \in \mathcal{M}$ three prediction heads recover the token $x_i$, terminal tag $\tau_i$, and non-terminal tag $\nu_i$, yielding $\mathcal{L} = \mathcal{L}_{\text{MLM}} + \mathcal{L}_{\text{T}} + \mathcal{L}_{\text{NT}}$; one masked position is shown, and head colors match the embedding rows they supervise. For decoders, next-token prediction replaces MLM: the token head at position $i$ predicts token $i{+}1$ at every position, while the tag heads are applied only at first-subword positions and predict the tags of the \emph{next tagged} position (\autoref{subword-tag-injection}).}
\label{fig:approach}
\end{figure*}

\input{sections/introduction}

\input{sections/related_work}

\input{sections/method}
\begin{table}[t]
\centering
\resizebox{\columnwidth}{!}{%
\begin{tabular}{l}
\toprule
\rowcolor{lightgray}
\textit{BLIMP Example} \\
\greencheck ~~The keys to the cabinet \textcolor{red}{\bf are} on the table. \\
\redcross ~~~The keys to the cabinet \textcolor{red}{\bf is} on the table. \\
\midrule
\rowcolor{lightgray}
\textit{SyntaxGym Example} \\
\greencheck ~~The farmer near the clerks \textcolor{red}{\bf knows} many people. \\
\redcross ~~~The farmer near the clerks \textcolor{red}{\bf know} many people. \\
\bottomrule
\end{tabular}
}
\caption{Examples from the BLIMP dataset (top) and SyntaxGym (bottom). In both datasets, $p($\greencheck$)> p($\redcross$)$, but in BLIMP this probability is computed for the complete sentence, whereas in SyntaxGym it is computed only over the region of interest.}
\label{tab:dataexamples}
\end{table}

\begin{table}[!t]
\centering
\small
\renewcommand{\arraystretch}{1.2}
\setlength{\tabcolsep}{1.5pt}
\resizebox{\columnwidth}{!}{%
\begin{tabular}{@{}p{68pt}cccccc@{}}
\toprule
\textbf{Model} &
\makecell{\textbf{BLiMP}\\$\uparrow$} &
\makecell{\textbf{$\Delta$}\\\textbf{BLiMP}} &
\makecell{\textbf{PPL}\\$\downarrow$} &
\makecell{\textbf{$\Delta$}\\\textbf{PPL}} &
\makecell{\textbf{Syntax}\\\textbf{Gym} $\uparrow$} &
\makecell{\textbf{$\Delta$}\\\textbf{SyntaxGym}} \\
\midrule
\sectrow{No Inductive Bias}
Transformer-XL (tokens) & 75.30 & \basedelta & 18.63 & \basedelta & 73.09 & \basedelta \\
GPT-2 (tokens) \cite{radford2019language} & 72.20 & \downdelta{-4.12} & 21.60 & \downdelta{-15.94} & 71.90 & \downdelta{-1.63} \\
\midrule
\sectrow{Constituency Parsing}
PLM (GPT-2) \cite{qian-etal-2021-structural} & 75.10 & \downdelta{-0.27} & 29.80 & \downdelta{-59.96} & 80.20 & \updelta{9.73} \\
Transformer Grammar (Transformer-XL) \cite{Sartran_2022} & 73.50 & \downdelta{-2.39} & 18.40 & \updelta{1.23} & \textbf{82.50} & \textbf{\updelta{12.87}} \\
Pushdown LM (GPT-2) \cite{murty2023pushdownlayersencodingrecursive} & \textbf{75.60} & \textbf{\updelta{0.40}} & 19.90 & \downdelta{-6.82} & 82.30 & \updelta{12.60} \\
TreeReg (GPT-2) \cite{nandi2025sneakingsyntaxtransformerlanguage} & 74.80 & \downdelta{-0.66} & 22.30 & \downdelta{-19.70} & 80.00 & \updelta{9.45} \\
\midrule
\sectrow{Tree-Planted Transformers (TPT) \cite{Yoshida_2024}}
TPT [zero] & --- & --- & 47.50 & \downdelta{-154.97} & 71.70 & \downdelta{-1.90} \\
TPT [rand.] & --- & --- & 47.40 & \downdelta{-154.43} & 69.00 & \downdelta{-5.60} \\
TPT [seq.] & --- & --- & 47.30 & \downdelta{-153.89} & 70.10 & \downdelta{-4.09} \\
TPT [dep.] & --- & --- & 47.70 & \downdelta{-156.04} & 77.10 & \updelta{5.49} \\
TPT [cons.] & --- & --- & 45.50 & \downdelta{-144.23} & 75.80 & \updelta{3.71} \\
TPT [bin.] & --- & --- & 45.60 & \downdelta{-144.77} & 73.00 & \downdelta{-0.12} \\
\midrule
\sectrow{Syntax-Informed Embeddings with Hexatagging (Ours)}
\multicolumn{7}{@{}l}{\textit{SiPE (positional pathway)}} \\
\quad Input-side Injection & 73.82 & \downdelta{-1.97} & 16.16 & \updelta{13.26} & 76.97 & \updelta{5.31} \\
\quad Position-pathway Injection & 74.01 & \downdelta{-1.71} & 16.95 & \updelta{9.02} & 80.60 & \updelta{10.29} \\
\quad Fully-disentangled Injection & 74.72 & \downdelta{-0.77} & 16.66 & \updelta{10.57} & 78.72 & \updelta{7.70} \\
\cdashline{1-7}
\multicolumn{7}{@{}l}{\textit{Attention Bias}} \\
\quad Projections shared across layers  & 73.60 & \downdelta{-2.26} & 16.73 & \updelta{10.20} & 76.35 & \updelta{4.46} \\
\quad Projections Learned Per-Layer & 74.66 & \downdelta{-0.85} & 16.24 & \updelta{12.83} & 78.10 & \updelta{6.86} \\
\cdashline{1-7}
\multicolumn{7}{@{}l}{\textit{SiPE + Attention Bias}} \\
\quad Input-side Injection & 74.16 & \downdelta{-1.51} & \textbf{15.89} & \textbf{\updelta{14.71}} & 78.47 & \updelta{7.36} \\
\quad Position-pathway Injection & 74.44 & \downdelta{-1.14} & 16.52 & \updelta{11.32} & 77.97 & \updelta{6.68} \\
\bottomrule
\end{tabular}%
}
\caption{\footnotesize Language modeling and syntax evaluation of decoder LMs; all our SiPE variants (Ours) use Transformer-XL as the decoder model, and each baseline's model is noted in parentheses. BLiMP and SyntaxGym are higher-is-better, while PPL (perplexity, computed on BLLIP-LG test split \cite{charniak2000bllip}) is lower-is-better. $\Delta$ columns report relative improvement or degradation against vanillaTransformer-XL (tokens). Best scores for each dataset are \textbf{bolded}.}
\label{tab:results_delta-dup1}
\end{table}

\begin{table}[!t]
\centering
\small
\setlength{\tabcolsep}{3.5pt}
\renewcommand{\arraystretch}{1.15}
\begin{tabularx}{\columnwidth}{@{}Xrr@{}}
\toprule
\textbf{Model} & \textbf{Acc. (\%)} & \textbf{$\Delta$ (\%)} \\
\midrule
\rowcolor{gray!15}
\multicolumn{3}{@{}c@{}}{\textit{Our models on WikiText}} \\
RoBERTa-base & 70.68 & -- \\
RoBERTa + SiPE  (Input pathway)& 71.68 & \textcolor{green}{$\uparrow$} +1.41 \\
DeBERTa-base & 67.40 & -- \\
DeBERTa + SiPE (Input pathway) & 68.93 & \textcolor{green}{$\uparrow$} +2.27 \\
ModernBERT-base & 49.94 & -- \\
ModernBERT + SiPE (Input pathway) & 51.10 & \textcolor{green}{$\uparrow$} \textbf{+2.32} \\
Transformer-XL & 53.56 & -- \\
Transformer-XL + SiPE (Positional Pathway) & 53.44 & \textcolor{red}{$\downarrow$} -0.22  \\
\addlinespace[2pt]
\rowcolor{gray!15}
\multicolumn{3}{@{}c@{}}{\textit{OOD Eval: WikiText $\to$ BLLIP-LG}} \\
RoBERTa-base & 71.29 & -- \\
RoBERTa + SiPE (Input pathway) & 72.62 & \textcolor{green}{$\uparrow$} +1.87 \\
DeBERTa-base & 69.01 & -- \\
DeBERTa-base + SiPE (Input pathway) & 71.12 & \textcolor{green}{$\uparrow$} +3.06 \\
ModernBERT-base & 51.25 & -- \\
ModernBERT-base + SiPE (Input pathway) & 53.41 & \textcolor{green}{$\uparrow$} \textbf{+4.21}  \\

\bottomrule
\end{tabularx}
\caption{\footnotesize{BLiMP accuracy results for models trained with SiPE.
For each model family we report its most performant GLUE variant: input-pathway
SiPE injection for the encoders and positional-pathway SiPE injection for the
Transformer-XL decoder; the OOD experiments likewise use the most performant
GLUE variant of each family.
For encoders, we use PLL-based accuracy.
$\Delta$: relative change w.r.t.\ the corresponding base model.
OOD section: models pre-trained on WikiText, then continuously pre-trained on
BLLIP-LG \cite{charniak2000bllip}.
\textcolor{green}{$\uparrow$} improvement, \textcolor{red}{$\downarrow$} degradation.
\textbf{Bold}: best relative improvement with our method.}}
\label{tab:blimp_comparison}
\end{table}

\begin{table*}[t]
\centering
\small
\setlength{\tabcolsep}{5pt}
\renewcommand{\arraystretch}{1.5}
\begin{tabular}{@{}lc>{\columncolor{sipecol}}cc>{\columncolor{sipecol}}c>{\columncolor{ppcol}}cc>{\columncolor{sipecol}}c>{\columncolor{ppcol}[\tabcolsep][0pt]}c@{}}
\toprule
& \multicolumn{2}{c}{\textbf{RoBERTa (base)}} & \multicolumn{3}{c}{\textbf{DeBERTa (base)}} & \multicolumn{3}{c}{\textbf{ModernBERT (base)}} \\
\cmidrule(lr){2-3} \cmidrule(lr){4-6} \cmidrule(l){7-9}
\textbf{Task}
& Base & +SiPE
& Base & +SiPE & +PP-SiPE
& Base & +SiPE & +PP-SiPE \\
\midrule
CoLA
& \res{31.41}{2.02} & \res{\textbf{31.77}}{0.71}
& \res{\textbf{27.22}}{2.37} & \res{26.09}{2.14} & \res{24.97}{2.02}
& \res{\textbf{25.00}}{0.83} & \res{24.45}{1.85} & \res{23.26}{2.83} \\
SST-2
& \res{\textbf{87.23}}{0.79} & \res{87.22}{0.52}
& \res{87.31}{0.30} & \res{\textbf{87.88}}{0.54} & \res{87.50}{0.19}
& \res{85.09}{1.01} & \res{\textbf{85.86}}{0.79} & \res{84.25}{1.38} \\
QQP
& \res{83.24}{0.21} & \res{\textbf{83.36}}{0.11}
& \res{84.02}{0.31} & \res{\textbf{84.95}}{0.15} & \res{84.88}{0.13}
& \res{\textbf{83.53}}{0.13} & \res{83.09}{0.45} & \res{82.45}{0.94} \\
QNLI
& \res{77.65}{0.50} & \res{\textbf{78.24}}{0.27}
& \res{68.07}{0.08} & \res{\textbf{68.88}}{0.24} & \res{67.94}{0.29}
& \res{62.51}{1.01} & \res{61.83}{0.70} & \res{\textbf{65.79}}{0.39} \\
MNLI
& \res{73.03}{0.33} & \res{\textbf{74.05}}{0.17}
& \res{71.25}{0.14} & \res{\textbf{71.85}}{0.14} & \res{71.74}{0.30}
& \res{71.29}{0.03} & \res{\textbf{72.24}}{0.35} & \res{71.15}{0.11} \\
RTE
& \res{\textbf{65.76}}{0.74} & \res{64.66}{0.56}
& \res{62.94}{1.23} & \res{63.06}{1.73} & \res{\textbf{64.14}}{1.68}
& \res{62.21}{0.45} & \res{63.06}{2.13} & \res{\textbf{63.90}}{1.02} \\
STSB
& \res{79.11}{0.30} & \res{\textbf{81.41}}{0.13}
& \res{76.20}{0.56} & \res{\textbf{76.98}}{0.24} & \res{76.33}{1.32}
& \res{79.67}{0.30} & \res{\textbf{81.21}}{0.78} & \res{79.49}{0.06} \\
MRPC
& \res{80.96}{0.44} & \res{\textbf{82.02}}{0.85}
& \res{80.49}{0.84} & \res{81.11}{0.76} & \res{\textbf{83.45}}{0.71}
& \res{79.74}{1.43} & \res{\textbf{82.82}}{1.08} & \res{81.56}{1.07} \\
\midrule
\textbf{Macro}
& \res{72.30}{0.33} & \res{\textbf{72.84}}{0.42}
& \res{69.69}{0.25} & \res{70.10}{0.39} & \res{\textbf{70.12}}{0.22}
& \res{68.63}{0.08} & \res{\textbf{69.32}}{0.11} & \res{68.98}{0.37} \\
\bottomrule
\end{tabular}
\caption{\footnotesize GLUE evaluation of baseline encoders vs.\ SiPE augmentation, over 3 seeds (std in parentheses). Baselines span three PE schemes: RoBERTa-base (absolute), DeBERTa-v3-base (relative), ModernBERT-base (RoPE). SiPE is injected via two routes: \textcolor{green!60!black}{input pathway} (into token embeddings) and \textcolor{blue!70!black}{positional pathway} (into positional information; relative/RoPE only). Best variant per task in \textbf{bold}. SiPE improves every architecture on the macro average---RoBERTa $+0.75\%$, DeBERTa-v3 $+0.59\%$, ModernBERT $+1.0\%$ relative.
}
\label{tab:glue_base_encoders}
\end{table*}
\begin{table}[t]
\centering
\small
\setlength{\tabcolsep}{4pt}
\renewcommand{\arraystretch}{1.5}
\begin{tabular}{@{}lc>{\columncolor{sipecol}}cc>{\columncolor{sipecol}[\tabcolsep][0pt]}c@{}}
\toprule
& \multicolumn{2}{c}{\textbf{RoBERTa (large)}} & \multicolumn{2}{c}{\textbf{ModernBERT (large)}} \\
\cmidrule(lr){2-3} \cmidrule(l){4-5}
\textbf{Task}
& Base & +SiPE
& Base & +SiPE \\
\midrule
CoLA
& \res{\textbf{30.71}}{0.51} & \res{30.36}{0.88}
& \res{\textbf{25.06}}{1.21} & \res{24.58}{0.87} \\
SST-2
& \res{\textbf{88.95}}{0.35} & \res{87.50}{0.09}
& \res{84.44}{0.52} & \res{\textbf{85.47}}{0.36} \\
QQP
& \res{84.31}{0.18} & \res{\textbf{85.10}}{0.24}
& \res{82.54}{0.39} & \res{\textbf{82.55}}{0.55} \\
QNLI
& \res{80.18}{0.18} & \res{\textbf{82.68}}{0.33}
& \res{61.99}{0.52} & \res{\textbf{62.11}}{0.41} \\
MNLI
& \res{\textbf{74.72}}{0.10} & \res{74.43}{0.31}
& \res{72.03}{0.53} & \res{\textbf{72.36}}{0.22} \\
RTE
& \res{\textbf{65.94}}{0.45} & \res{64.62}{1.06}
& \res{\textbf{63.54}}{0.29} & \res{61.49}{2.27} \\
STSB
& \res{81.89}{0.59} & \res{\textbf{82.34}}{0.38}
& \res{79.03}{0.41} & \res{\textbf{80.05}}{0.35} \\
MRPC
& \res{80.92}{1.00} & \res{\textbf{83.70}}{1.15}
& \res{81.86}{1.24} & \res{\textbf{82.00}}{0.76} \\
\midrule
\textbf{Macro}
& \res{73.45}{0.16} & \res{\textbf{73.84}}{0.13}
& \res{68.81}{0.13} & \res{\textbf{68.83}}{0.35} \\
\bottomrule
\end{tabular}
\\[2pt]
{\footnotesize\raggedright We did not pre-train DeBERTa-large with SiPE (input pathway/positional pathway), nor ModernBERT-large with SiPE (positional pathway), due to limited academic compute and budget constraints.\par}
\caption{\footnotesize GLUE evaluation results for large encoder models.}


\label{tab:glue_large_encoders}
\end{table}

\begin{table}[t]
\centering
\small
\setlength{\tabcolsep}{10pt}
\renewcommand{\arraystretch}{1.5}
\begin{tabular}{@{}lc>{\columncolor{ppcol}[\tabcolsep][0pt]}c@{}}
\toprule
& \multicolumn{2}{c}{\textbf{Transformer-XL}} \\
\cmidrule(l){2-3}
\textbf{Task}
& Base
& +PP-SiPE \\
\midrule
CoLA  & \res{13.80}{0.26} & \res{\textbf{26.26}}{0.22} \\
SST-2 & \res{84.14}{0.11} & \res{\textbf{88.80}}{0.44} \\
QQP   & \res{83.83}{0.12} & \res{\textbf{85.64}}{0.21} \\
QNLI  & \res{76.00}{0.26} & \res{\textbf{82.28}}{0.29} \\
MNLI  & \res{68.07}{0.32} & \res{\textbf{75.22}}{0.14} \\
RTE   & \res{63.42}{1.70} & \res{\textbf{67.51}}{0.29} \\
STSB  & \res{76.90}{0.17} & \res{\textbf{81.80}}{0.12} \\
MRPC  & \res{79.20}{0.21} & \res{\textbf{82.75}}{0.26} \\
\midrule
\textbf{Macro}
& \res{68.17}{0.17}
& \res{\textbf{73.78}}{0.05} \\
\bottomrule
\end{tabular}
\caption{\footnotesize GLUE results for the decoder-only Transformer-XL (TXL)
baseline and its SiPE (positional pathway) variant
(\textcolor{blue!70!black}{blue tint}); best per task in bold. SiPE improves
every task and gains +8.2\% relative on the macro average
(68.17$\rightarrow$73.78), showing the positional pathway best augments TXL's
relative positional encoding.}
\label{tab:glue_decoder_txl}
\end{table}

\input{sections/results}

\input{sections/layerwise}

\input{sections/conclusion}
\section*{Limitations}
Our approach assumes access to hexatag annotations at inference time: each input
sequence must be tagged by a dependency parser before being passed to the
syntax-infused encoder. While this adds a preprocessing step, the tagger is
lightweight and fast in practice (we use \texttt{DeBERTa-v3-base} as the
hexatagger, 184M parameters) and runs efficiently relative to encoder
pre-training and downstream finetuning. The approach also inherits the hexatagger's
errors, so its benefit may be smaller in domains where the parser is less
accurate.

We consider only coarse-grained syntactic priors derived from directional
terminal and non-terminal tags. Preliminary pre-training experiments
incorporating full dependency-relation labels (\textsc{deprel}) did not yield
meaningful gains on GLUE (Appendix~\ref{app:alternate-combination-appendix},
Figures~\ref{fig:glue_avg_ablation} and~\ref{fig:deprel_per_dataset}), but we did
not investigate this further.

Our study of where to inject the prior is broad but not exhaustive. We study the
main injection sites and coupling choices for each positional scheme, but under a
fixed academic compute budget we could not pre-train every conceivable
combination, particularly for the relative and rotary attention-side variants,
where the design space is large. The configurations we report are therefore the
strongest we found rather than a guaranteed global optimum, and our per-dataset
results with RoBERTa (Figure~\ref{fig:deprel_per_dataset}) show that no single setting
dominates across all tasks.

Perhaps the biggest limitation, which we plan to address in future work, is
fast autoregressive text generation with the decoder. Because the injected prior
is conditioned on hexatags, generating each new token requires re-tagging the
sequence produced so far. Our hexatagger is fast, so this is feasible in
principle, but it is not compatible with standard KV-caching: as the sequence
grows, the parser may revise the tags of earlier tokens, changing their injected
representations and invalidating the cached keys and values. Efficient
incremental decoding under a per-step syntactic prior is therefore an open
research-engineering problem, and the direction we consider most important for
future work.

Finally, our experiments are limited to small models, English text, and
the pre-training budgets used here. Whether the gains hold at larger scale, in
other languages, or under substantially longer pre-training, especially with billion parameter LLMs remains to be seen; its unclear how the
relative benefit of an explicit syntactic prior will scale as model and data scale increase.



\bibliography{custom}

\input{sections/appendix}

\end{document}

%% file: sections/introduction.tex
\section{Introduction}
\label{sec:intro}
The transformer architecture \cite{vaswani2017attention} forms the foundation of modern LLMs. While the impact of LLMs is substantial, several linguistic simplifications were introduced in the underlying architecture to improve scalability.
One key choice was the use of positional embeddings (PE) to encode each token's location within a sequence. While this approach preserved computational efficiency and scalability, it reduced positional information to a simple ordering mechanism—sufficient to indicate \emph{where} tokens appear, but insufficient to capture the \emph{syntactic relations} that form the scaffolding for downstream semantics. 
That is, without explicitly modeling syntactic compositionality, semantic frames~\cite{baker1998berkeley,kingsbury2002treebank} are more likely to be extracted incorrectly, increasing the risk of misinterpretation for LLM-based agents.
Consider, for example, this hypothetical agent request: {\em ``Please move this large file to another folder.’’} To execute the requested action correctly, the {\em move} predicate must be associated with its arguments: the object to be moved ({\em file}) and the destination ({\em folder}). Absolute positional encodings~\cite{devlin2019bertpretrainingdeepbidirectional,liu2019robertarobustlyoptimizedbert}, provide no signal to link this predicate to its arguments. Relative positional encodings, e.g., DeBERTa~\cite{he2021debertadecodingenhancedbertdisentangled} and Transformer-XL~\cite{dai2019transformerxlattentivelanguagemodels}, weaken this signal. For instance, the relative distance between {\em move} and {\em folder} is large (six words), even though a direct {\tt oblique} syntactic dependency connects the verb to its modifier~\cite{de2021universal}.

Partially due to this shallower representation, language models (LMs) still exhibit weak compositional generalization \cite{guo2020hierarchical}.
Prior work injects hierarchical inductive bias by constraining self-attention using constituency or dependency structure, often via quadratic token-token interactions and hard-coded attention masks \cite{Sartran_2022, zhao2024dependencytransformergrammarsintegrating, murty2023pushdownlayersencodingrecursive, qian-etal-2021-structural, xie-etal-2021-transformer}. 
This is a complex process that requires considerable changes in the transformer
architecture and which increases the attention mechanism's computational
overhead (e.g., Pushdown LMs~\cite{murty2023pushdownlayersencodingrecursive}
maintain a recursively updated stack over the sequence that reshapes the
attention pattern at every step).

The main contributions of our paper are:

{\flushleft {\bf (1)}} We introduce \textbf{S}yntax-\textbf{i}nformed \textbf{P}ositional \textbf{E}mbeddings (\textbf{SiPE}) that augment a model's positional pathway with a lightweight prior derived from linearized dependency trees---coarse directional indicators relating each token to its syntactic governor (\S\ref{sec:approach}), learned during pre-training via an auxiliary 
indicator-prediction
objective alongside the LM loss.

{\flushleft {\bf (2)}} SiPE composes with all three dominant positional encoding families---absolute~\cite{liu2019robertarobustlyoptimizedbert}, relative~\cite{he2021debertadecodingenhancedbertdisentangled,dai2019transformerxlattentivelanguagemodels}, and rotary~\cite{warner2024smarterbetterfasterlonger}---without modifying self-attention or any other Transformer component.

{\flushleft {\bf (3)}} For encoders, SiPE improves BLiMP \cite{warstadt2023blimpbenchmarklinguisticminimal} syntactic generalization across all three families (up to $+2.3\%$ relative for ModernBERT), with gains compounding under continued pre-training on BLLIP-LG ($+3\%$ for DeBERTa-v3 on out-of-domain evaluation), indicating the prior generalizes beyond its pre-training distribution (\autoref{tab:blimp_comparison}). The same models also improve on GLUE~\cite{wang2019gluemultitaskbenchmarkanalysis} at both small and large scale (tables \autoref{tab:glue_base_encoders},  \ref{tab:glue_large_encoders} and \ref{tab:glue_decoder_txl}).


{\flushleft {\bf (4)}} For autoregressive decoders like Transformer-XL, injecting through the positional pathway alone is the strongest recipe: it achieves the best SyntaxGym \cite{hu2020systematicassessmentsyntacticgeneralization} score of any configuration of our method ($80.60$, $+10.3$ relative improvement over the no-syntax baseline) while cutting perplexity by $9.0\%$ (relative), and drives a significant $+8.2\%$ relative improvement on GLUE for the same model (\autoref{tab:glue_decoder_txl}).


{\flushleft {\bf (5)}} As shown in  \autoref{fig:pareto}, prior syntactic LMs sit at two extremes: parsing at inference (expensive marginalization over many parse trees) or only during training (discarding syntax at runtime, with weaker syntactic generalization). SiPE moves the Pareto frontier between these by conditioning on a \emph{single} parse at inference. 
The injection of syntactic information adds no asymptotic cost over self-attention,\footnote{\label{foot:parser}Dependency tags are predicted in parallel by a lightweight \texttt{DeBERTa-v3-base} classifier whose overhead on top of the encoder's is linear in input length; see \autoref{sec:time-complexity}. Implemented in-house with a multitask-learning encoder, two per-token tag heads: \url{https://clulab.org/processors/}} and unlike most prior syntactic LMs which jointly model syntax and language, SiPE leaves downstream usage unchanged---models architecturally remain compatible with continual pre-training, supervised fine-tuning and reinforcement learning. Empirically we outperform parser-free approaches---Tree-Planted Transformers~\cite{Yoshida_2024} and TreeReg~\cite{nandi2025sneakingsyntaxtransformerlanguage}---on both BLLIP-LG perplexity and SyntaxGym (\autoref{tab:results_delta-dup1}).

%% file: sections/related_work.tex
\section{Related Work, Background, and Notations}
\label{sec:related-work}
Prior work shows that pretraining on formal languages or injecting tree-based signals can improve syntactic generalization and data efficiency \cite{hu2025circuitschomskyprepretrainingformal}.
However, most methods encode syntax by constraining self-attention via pairwise token interactions and hard-coded attention masks,
adding parameters and training complexity \cite{xie-etal-2021-transformer, Sartran_2022, qian-etal-2021-structural, murty2023pushdownlayersencodingrecursive}.
We instead sidestep this by encoding dependency structure as two lightweight prior vectors derived from \textit{Hexatagging}, and apply these priors to the positional embedding pathway. Modern Transformers encode position through three dominant schemes: \emph{absolute} embeddings added to the input residual~\cite{liu2019robertarobustlyoptimizedbert}, \emph{relative} embeddings indexed by offset $i - j$ inside attention~\cite{he2023debertav3improvingdebertausing,dai2019transformerxlattentivelanguagemodels}, and \emph{rotary} embeddings that rotate queries and keys by position-dependent angles~\cite{su2023roformerenhancedtransformerrotary}. We show that our method seamlessly composes with all three PE types (\autoref{sec:input_injection}).
\paragraph{Hexatagging.}
In this work, we leverage \emph{Hexatagger} \cite{amini2023hexataggingprojectivedependencyparsing}, a ``parsing-as-tagging'' dependency parser that assigns each token two types of discrete syntactic labels (analogous to position or subword IDs): (a) the position of its syntactic governor (which can further be decomposed into two subtypes) and (b) the dependency relation. This formulation can be mapped to simple embedding lookup tables that can be linearly combined with token embeddings without modifying self-attention.
Hexatagger maps a projective dependency tree to a \emph{binary head tree} (BHT, a special form of constituency tree) by binarizing the structure and labeling each internal node with $L/R$ to indicate whether the span head lies in the left or right subtree, then linearizes the BHT via an in-order (left-corner) traversal.
Each visited node yields a tag encoding its attachment direction (left vs.\ right child), and for non-terminals, the head-direction ($L$ vs.\ $R$). 
This yields a fixed projective tag inventory with terminal tags
$\leftchild$ and $\rightchild$, plus non-terminal tags
$\Leftchildl$, $\Leftchildr$, $\Rightchildl$, and $\Rightchildr$
(\autoref{fig:approach}, left).
This avoids unbounded (length-dependent) tag sets whose cardinality grows with the input length; instead each token's tag can be predicted independently with a linear classifier. In our implementation, we further append a dedicated $\mathsf{EOS}$ non-terminal (via a left shift of between-token non-terminals) to obtain one non-terminal label per token.\footref{foot:parser} In this work, we omit dependency relation labels, finding that terminal/non-terminal hexatags alone provide sufficient syntactic inductive bias to improve downstream language modeling performance.

%% file: sections/method.tex
\section{Approach}

\label{sec:approach}
Both training and inference operate on a hexatagged sequence: the input is first
tagged by our parser (\autoref{subword-tag-injection}). Encoders train with the
standard MLM objective; decoder training is standard next-token prediction, as
each token carries its own tags and truncating the sequence at any prefix leaves
the retained tags intact.

\subsection{Deriving a Positional Syntactic Bias}
\label{sec:input_injection}

An inductive bias can enter the PE layer in three ways
(\autoref{fig:approach}):
(a) \emph{entangled with the token embedding}, applied to the input before
any positional or attention information is introduced; (b) \emph{composed
with the default PE scheme} under some operation; or (c) \emph{disentangled},
a separate additive term in the attention score that modulates each head.
Linearizing the binary head tree into discrete tags (the \emph{hexatagging}
procedure of \autoref{sec:related-work}) yields two embedding tables, one per
tag type, composable with the PE mechanism under any of these operations. At
position $p$ they contribute:
\begin{equation}
m_p \cdot \bigl( \textcolor{tagcolor}{\mathbf{E}^{T}_{\tau_p}} + \textcolor{tagcolor}{\mathbf{E}^{N}_{\nu_p}} \bigr)
\label{eq:tag_term}
\end{equation}
where $\textcolor{tagcolor}{\mathbf{E}^{T}} \in \mathbb{R}^{|\mathcal{T}| \times d}$
and $\textcolor{tagcolor}{\mathbf{E}^{N}} \in \mathbb{R}^{|\mathcal{N}| \times d}$
are learned tables ($|\mathcal{T}|{=}2$ terminals, $|\mathcal{N}|{=}5$
non-terminals), and $m_p \in \{0,1\}$ is the first-subword mask, equal to 1
only at each word's first subword so the hexatag is injected once per word
(\autoref{subword-tag-injection}). Tables are randomly initialized and trained jointly with the LM objective.
Throughout the paper, every occurrence of
$\textcolor{tagcolor}{\mathbf{E}^{T}_{\tau_p}} + \textcolor{tagcolor}{\mathbf{E}^{N}_{\nu_p}}$
is implicitly multiplied by $m_p$; we omit the mask from later equations for
readability, so continuation subwords contribute no tag term anywhere.

\subsection{RoBERTa: Absolute Positional Embeddings}
\label{sec:roberta_absolute}

We begin with the simplest case. Adopting strategy~(a), hexatag priors are
injected directly into the input embedding, which combines a learned absolute position
vector $\mathbf{p}_p$ with the token, segment, and tag terms:
\begin{equation}
\mathbf{x}_p = \mathbf{e}^{\text{tok}}_p + \mathbf{p}_p + \mathbf{e}^{\text{seg}}_p + m_p \cdot \bigl( \textcolor{tagcolor}{\mathbf{E}^{T}_{\tau_p}} + \textcolor{tagcolor}{\mathbf{E}^{N}_{\nu_p}} \bigr)
\label{eq:roberta_input}
\end{equation}
The tag term sits at the same level as $\mathbf{p}_p$, so the model learns to attend to a richer composite input without architectural conflict\footnote{However, the input embedding is not the only place in the transformer where we can inject these priors; see appendix \ref{app:alternate-combination-appendix} and figure \ref{fig:ablation}.}.

\subsection{Transformer-XL: Syntactic Priors in an Autoregressive Decoder with Relative PE}
\label{sec:txl_pebias}
To test prior injection on autoregressive decoders, we use
Transformer-XL~\cite{dai2019transformerxlattentivelanguagemodels} (hereon referred to as TXL), which is one of the earliest ``LLM-like'' architectures carrying a sinusoidal relative-position encoding directly in the attention
score and is a common syntactic-LM
baseline~\cite{Sartran_2022,zhao2024dependencytransformergrammarsintegrating}.
Following prior work, we disable cross-segment caching (so relative positions
apply only within the current window) and replace the adaptive softmax with a
tied linear projection, making TXL a purely causal decoder.
For a query at position $i$ and key at position $j$ in head $n$ of layer
$\ell$, the attention score splits into a content term and a position term:
%
\begin{equation}
\resizebox{\columnwidth}{!}{$\displaystyle
A^{(\ell)}_{i,j,n}
\;=\;
\underbrace{\bigl\langle \mathbf{q}^{(\ell)}_{i,n} + \mathbf{u}_n,\; \mathbf{k}^{(\ell)}_{j,n}\bigr\rangle}_{\textstyle \mathbf{AC}^{(\ell)}_{i,j,n}\ \text{(content)}}
\;+\;
\underbrace{\bigl\langle \mathbf{q}^{(\ell)}_{i,n} + \mathbf{v}_n,\; \mathbf{W}_R\, \textcolor{relposcolor}{\mathbf{r}_{i-j,n}}\bigr\rangle}_{\textstyle \textcolor{relposcolor}{\mathbf{BD}^{(\ell)}_{i,j,n}}\ \text{(position)}}
$}
\label{eq:txl_acbd}
\end{equation}
where $\textcolor{relposcolor}{\mathbf{r}_{i-j,n}}$ is the sinusoidal vector for
offset $i\!-\!j$, $\mathbf{W}_R$ projects it into head $n$'s
$d_{\text{head}}$-dimensional subspace, and $\mathbf{u}_n,\mathbf{v}_n$ are the
learned content and position bias vectors. Notably, this content-conditioned
relative-position term remains competitive at frontier scale: the concurrent
975B-parameter open-weights model Inkling~\cite{thinkingmachines2026inkling}
abandons RoPE for a relative positional bias of the same
query--offset-embedding form, citing better quality and length extrapolation.
Because
$\textcolor{relposcolor}{\mathbf{BD}}$ depends only on $i\!-\!j$, it carries no
information about \emph{which} tokens occupy positions $i$ and $j$ or their
syntactic role. We supply that missing signal from the hexatag embeddings
$\mathbf{E}^T_{\tau_j} + \mathbf{E}^N_{\nu_j}$ and study where in the score it
should enter, across five injection sites treated in turn below.

\subsubsection{Input-Side Injection (ADD)}
\label{sssec:txl_add}
The simplest approach adds the tag embeddings to the token embedding, so the
prior enters the input residual at layer $0$ and reaches the score only
indirectly, through the content projections $\mathbf{W}_Q,\mathbf{W}_K$ that
build the query and key:
%
\begin{equation}
\mathbf{h}^{(0)}_p \;=\; \mathbf{e}^{\text{tok}}_p \;+\; m_p\,(\textcolor{tagcolor}{\mathbf{E}^T_{\tau_p}} + \textcolor{tagcolor}{\mathbf{E}^N_{\nu_p}})
\label{eq:txl_add}
\end{equation}
with $m_p\!\in\!\{0,1\}$ masking tag-free positions. The relative-position
encoding itself is untouched; the prior entangles with lexical content rather than with position.


\subsubsection{Position-Pathway Injection (PP-SiPE)}
\label{sssec:txl_entangled}

The approach which we find works best is to leave the input residual clean and route the tag straight into the
position term. Using a layer-specific projection $\textcolor{tagcolor}{W_E^{(\ell)}}$, we map the
tag at key $j$ into head $n$'s subspace and take its inner product with the
\emph{same} effective query $\mathbf{q}^{(\ell)}_{i,n}+\mathbf{v}_n$ that appears
in the position term $\textcolor{relposcolor}{\mathbf{BD}}$ (Eq.~\ref{eq:txl_acbd}):
\begin{equation}
\resizebox{\columnwidth}{!}{$\displaystyle
\textcolor{gatecolor}{c^{(\ell)}_{i,j,n}}
\;=\; \tfrac{1}{\sqrt{d_{\text{head}}}}\,
\bigl\langle\, \mathbf{q}^{(\ell)}_{i,n} + \mathbf{v}_n,\;
\bigl[\textcolor{tagcolor}{W_E^{(\ell)}}(\textcolor{tagcolor}{\mathbf{E}^T_{\tau_j}} + \textcolor{tagcolor}{\mathbf{E}^N_{\nu_j}})\bigr]_n \,\bigr\rangle
$}
\label{eq:txl_sipe_c}
\end{equation}
where $[\,\cdot\,]_n$ selects head $n$'s $d_{\text{head}}$-dimensional block and where $\textcolor{tagcolor}{\mathbf{E}^T_{\tau_j}}$ and
$\textcolor{tagcolor}{\mathbf{E}^N_{\nu_j}}$ are the terminal and nonterminal
hexatag embeddings at key $j$.
Eqs.~\ref{eq:txl_acbd} and~\ref{eq:txl_sipe_c} are deliberately parallel: the
\emph{one} effective query $\mathbf{q}^{(\ell)}_{i,n}+\mathbf{v}_n$ poses two
questions --- ``how well do I align with this \emph{offset}?''
($\textcolor{relposcolor}{\mathbf{BD}}$) and ``how well do I align with this
\emph{tag}?'' ($\textcolor{gatecolor}{c}$). The second is therefore a
syntactic counterpart of the first, on the same per-head scale. We couple it to the position term multiplicatively, leaving the content term
$\mathbf{AC}$ untouched:
%
%
\begin{equation}
\fbox{$\displaystyle
\widetilde{A}^{(\ell)}_{i,j,n}
\;=\; \mathbf{AC}^{(\ell)}_{i,j,n}
\;+\; \bigl(1+\textcolor{gatecolor}{c^{(\ell)}_{i,j,n}}\bigr)\cdot \textcolor{relposcolor}{\mathbf{BD}^{(\ell)}_{i,j,n}}
$}
\label{eq:txl_sipe_mul}
\end{equation}
In our syntactic evaluation, we observe the multiplicative form (Eq.~\ref{eq:txl_sipe_mul}) to outperform all other approaches (refer to experiments in table \ref{tab:results_delta-dup1} and ablations in table \ref{tab:txl_pebias_ablation}, appendix \ref{app:txl_pebias_design}): its correction term
$\textcolor{gatecolor}{c}\cdot\textcolor{relposcolor}{\mathbf{BD}}$
\emph{gates by offset alignment}, applying the syntactic adjustment in
proportion to how strongly the query already attends to that offset. Where the
query-offset alignment is strong ($|\textcolor{relposcolor}{\mathbf{BD}}|$
large), the tag exerts a large pull; where it is near zero, the tag exerts
almost none, so syntax modulates an existing positional preference rather than
acting in isolation.\footnote{We initialize
$\textcolor{tagcolor}{W_E^{(\ell)}}$ small, so $\textcolor{gatecolor}{c}\!\approx\!0$ and
$\widetilde{A}\!\approx\!A$ at initialization: training starts from the vanilla
Transformer-XL score and learns the syntactic correction into it.} The offset embedding
$\textcolor{relposcolor}{\mathbf{r}_{i-j,n}}$ itself is never modified.

\subsubsection{Fully-Disentangled Injection}
\label{sssec:txl_disentangled}
A third option completely disentangles the tag from \emph{both} pathways. Rather than multiplying
$\textcolor{gatecolor}{c}$ with $\textcolor{relposcolor}{\mathbf{BD}}$, we add
it as an independent third component of the logit, alongside the content and
position terms and on the standard attention scale:
\begin{equation}
\widetilde{A}^{(\ell)}_{i,j,n}
\;=\; \mathbf{AC}^{(\ell)}_{i,j,n}
\;+\; \textcolor{relposcolor}{\mathbf{BD}^{(\ell)}_{i,j,n}}
\;+\; \textcolor{gatecolor}{c^{(\ell)}_{i,j,n}}.
\label{eq:txl_sipe_dis}
\end{equation}

Now $\textcolor{gatecolor}{c^{(\ell)}_{i,j,n}}$ depends on the query and the
tag but not on $\textcolor{relposcolor}{\mathbf{BD}^{(\ell)}_{i,j,n}}$, so its
size is fixed once the tag is chosen: a query at $i$ adds the same syntactic
bias toward key $j$ whether or not it already attends to the offset $i\!-\!j$.
This is the opposite of the multiplicative coupling
(Eq.~\ref{eq:txl_sipe_mul}), where that same bias is scaled by
$\textcolor{relposcolor}{\mathbf{BD}}$, and therefore drops to nearly zero when
the query does not attend to the offset (i.e., when
$\textcolor{relposcolor}{\mathbf{BD}}\!\approx\!0$). In our experiments, we find that this ungated form is weaker
(refer to Table~\ref{tab:results_delta-dup1} and ablations in Table \ref{tab:txl_disentangled_ablation}, appendix \ref{app:txl_disentangled_design}) than the multiplicatively entangled bias, which indicates that the bias from SiPE
helps most when it augments an attention preference the position pathway has
already established, rather than when it acts on its own.

\subsubsection{Injecting Syntactic Bias in Self-Attention}
\label{sssec:txl_attn}
Motivated by nearly all prior work which imposes syntactic constraints on attention weights as either a hard or soft bias, \cite{strubell-etal-2018-linguistically, murty2023pushdownlayersencodingrecursive, zhao2024dependencytransformergrammarsintegrating, omote2019dependencyrelativepos, xie-etal-2021-transformer, Yoshida_2024}, we propose our hexatag-derived bias applied to attention logits and the attention output, leaving the input residual, $\mathbf{AC}$, and the positional pathway $\textcolor{relposcolor}{\mathbf{BD}}$ untouched ($\mathbf{h}^{(0)}_p = \mathbf{e}^{\text{tok}}_p$).

The mechanism follows ~\citet{shaw2018selfattentionrelativepositionrepresentations}, who augment
attention with learned key-side and value-side vectors. We project the summed
hexatag embedding at key $j$ two ways: through $\mathbf{W}^{\text{tag},(\ell)}_K$
to form a key bias that is added to the attention score
$\widetilde{A}^{(\ell)}_{i,j,n}$, and through $\mathbf{W}^{\text{tag},(\ell)}_V$
to form a value bias that is added to the attended output
$\mathbf{z}^{(\ell)}_{i,n}$:
%
%
\begin{equation}
\resizebox{\columnwidth}{!}{$\displaystyle
\widetilde{A}^{(\ell)}_{i,j,n}
\;=\; A^{(\ell)}_{i,j,n}
\;+\; \tfrac{1}{\sqrt{d_{\text{head}}}}\bigl\langle \mathbf{q}^{(\ell)}_{i,n},\;
   [\mathbf{W}^{\text{tag},(\ell)}_K(\textcolor{tagcolor}{\mathbf{E}^T_{\tau_j}} + \textcolor{tagcolor}{\mathbf{E}^N_{\nu_j}})]_n \bigr\rangle
$}
\label{eq:txl_attn_k}
\end{equation}

\begin{equation}
\resizebox{\columnwidth}{!}{$\displaystyle
\mathbf{z}^{(\ell)}_{i,n}
\;=\; \sum_j \alpha^{(\ell)}_{i,j,n}\Bigl(\mathbf{v}^{(\ell)}_{j,n}
\;+\; [\mathbf{W}^{\text{tag},(\ell)}_V(\textcolor{tagcolor}{\mathbf{E}^T_{\tau_j}} + \textcolor{tagcolor}{\mathbf{E}^N_{\nu_j}})]_n\Bigr)
$}
\label{eq:txl_attn_v}
\end{equation}
where $\alpha^{(\ell)}_{i,j,n}$ are the softmax
weights, and $[\,\cdot\,]_n$ selects multi-head attention head $n$'s block. Two
things differ from the multiplicative alignment score $\textcolor{gatecolor}{c^{(\ell)}_{i,j,n}}$
of Eq.~\ref{eq:txl_sipe_c}. First, that score was built from the effective query
$\mathbf{q}^{(\ell)}_{i,n}+\mathbf{v}_n$ and scaled the position term
$\textcolor{relposcolor}{\mathbf{BD}}$, whereas this key bias uses the plain query $\mathbf{q}^{(\ell)}_{i,n}$ and
is added directly to the logit. Second, it is keyed on the tag at the single key
position $j$ rather than on a relative pair $(i,j)$, which keeps the added cost
linear (rather than quadratic) in sequence length. The projections may be learned per layer (the $(\ell)$
superscript) or tied to a single pair shared across all layers; in our
experiments we find the former (per-layer variant) outperforms the latter
(shared variant) on syntactic generalization tasks, both on its own and when
combined with the input-side injection of Eq.~\ref{eq:txl_add}
(Appendix~\ref{app:txl_attn_design}).

\subsubsection{Injecting Syntax into Position and Attention Simultaneously}
\label{sssec:txl_combined}

The final variant pairs our two strongest single-site injections: SiPE on the
positional pathway (the multiplicative coupling, Eq.~\ref{eq:txl_sipe_mul}) and the
per-layer attention-side biases. Combining the two, however, hurts: the joint variant scores below either
injection strategy used on its own
(Table~\ref{tab:results_delta-dup1}, Appendix~\ref{app:txl_attn_design}).
\textit{Injecting the same hexatag signal as both a positional and attentional bias is thus redundant rather
than complementary}; for Transformer-XL, the prior is best supplied once, through the positional
pathway. Having established \emph{where} and \emph{how} in the score the prior should be
combined, we next ask \emph{which layer} it should enter from: at least for the
decoder model in our experiments, {\bf syntax is best infused from layer~1
onward} (section \ref{sec:layerwise}).

\subsection{DeBERTa-v3: Injecting the Prior under Disentangled Relative Attention}
\label{sec:deberta}

DeBERTa-v3~\cite{he2023debertav3improvingdebertausing} carries no input-level
position embedding; positional information enters only at the attention score,
through \emph{disentangled relative attention}. For a query at $i$ and key at
$j$, the score splits into a content term and two relative-position terms:
\begin{equation}
S_{ij} \;=\;
\underbrace{Q^{c}_{i}{K^{c}_{j}}^{\!\top}}_{\textstyle S^{\text{con}}_{ij}}
\;+\;
\underbrace{Q^{c}_{i}{K^{r}_{\delta(i,j)}}^{\!\top}
        + K^{c}_{j}{Q^{r}_{\delta(i,j)}}^{\!\top}}_{\textstyle S^{\text{pos}}_{ij}} ,
\label{eq:deberta_score}
\end{equation}
where $\delta(i,j)$ is the bucketed relative offset and $Q^c,K^c$ are the
content query and key. As with the relative-PE models above, we consider two
injection sites: the input residual, and the position term $S^{\text{pos}}$.

\subsubsection{Input-Side Injection}
\label{sssec:deberta_add}
Because DeBERTa has no absolute position embedding, the tag prior becomes the
only positionally-localized signal at the input layer:
\begin{equation}
\mathbf{h}^{(0)}_p = \mathbf{e}^{\text{tok}}_p + \mathbf{e}^{\text{seg}}_p
   + m_p \!\cdot\! \bigl( \textcolor{tagcolor}{\mathbf{E}^{T}_{\tau_p}} + \textcolor{tagcolor}{\mathbf{E}^{N}_{\nu_p}} \bigr).
\label{eq:deberta_input}
\end{equation}
%
The prior propagates through $\mathbf{W}_Q$ and $\mathbf{W}_K$ into the
content terms of the attention score, while the relative-position mechanisms (its
bucketing, lookup table, and projection matrices) are left untouched. Tag priors
and relative position thus interact \emph{only through the content stream},
which is what makes the injection portable across positional-encoding families.
More details, including how the prior propagates to all three score terms, is given in Appendix~\ref{app:deberta_injection}.

\subsubsection{Position-Pathway Injection (PP-SiPE)}
\label{sssec:deberta_pebias}
Mirroring our Transformer-XL injection, we do not add the prior to the input
embeddings (the token representations enter the network unchanged) and instead
route it only into the relative-position term $S^{\text{pos}}_{ij}$ of the
attention score, leaving the content term $S^{\text{con}}_{ij}$ untouched. A per-layer projection
$W_{E}^{(\ell)}$ maps the summed hexatag embedding at key $j$ into head space,
and we form a tag--query alignment against the content query:
\begin{align}
c_{ij} &\;=\; \tfrac{1}{\sqrt{d_{h}}}\,
\big\langle Q^{c}_{i},\; W_{E}^{(\ell)}(\textcolor{tagcolor}{\mathbf{E}^T_{\tau_j}} + \textcolor{tagcolor}{\mathbf{E}^N_{\nu_j}})\big\rangle, \label{eq:deberta_c}\\[2pt]
\tilde{S}_{ij} &\;=\; S^{\text{con}}_{ij} \;+\; (1+c_{ij})\, S^{\text{pos}}_{ij}.
\label{eq:deberta_gate}
\end{align}
The coefficient $c_{ij}$ scales the relative-position score in proportion to
how strongly the query aligns with the key's syntactic tag, so syntax sharpens an
existing positional preference rather than acting on its own, exactly as in the
multiplicative Transformer-XL variant
(Eq.~\ref{eq:txl_sipe_mul}).\footnote{$W_{E}^{(\ell)}$ is initialized small so
that $c_{ij}\!\approx\!0$ and $\tilde{S}_{ij}\!\approx\!S_{ij}$ at
initialization; training starts from the unmodified DeBERTa score and learns the
syntactic correction into it.}


\subsection{ModernBERT: Injecting the Prior into Rotary Position Encoding}
\label{sec:modernbert}

ModernBERT~\cite{warner2024smarterbetterfasterlonger} like many other Large language models \cite{Dubey2024TheL3, deepseekai2024deepseekv3technicalreport, qwen3, Kamath2025Gemma3T} delivers positional
information through rotary positional embeddings
(RoPE)~\cite{su2023roformerenhancedtransformerrotary}, which rotate the content
query and key by position-dependent angles \emph{after} the content
projections, so that relative-position dependence emerges from the geometry of
two absolute rotations. Each head dimension is split into frequency pairs
indexed by $t$; pair $t$ is rotated by $\theta_{p,t}=p\,\omega_t$ at absolute
position $p$, with frequency $\omega_t$ and a $2{\times}2$ rotation $R(\cdot)$.
Because RoPE carries no input-level position term, we again consider two
injection sites: the input residual, and the rotation angle itself.

\subsubsection{Input-Side Injection}
\label{sssec:modernbert_add}
Since RoPE has no input-level position embedding, the tag prior is added at the
input residual exactly as in DeBERTa (Eq.~\ref{eq:deberta_input}, without the
segment term, which ModernBERT does not use), and propagates into
$\mathbf{q}_p,\mathbf{k}_p$ before rotation; the rotary mechanism itself is
untouched. 

\subsubsection{Position-Pathway Injection (PP-SiPE)}
\label{sssec:modernbert_pebias}
Mirroring the Transformer-XL and DeBERTa injections, we can also route the prior into the position pathway instead of the input embedding, which for RoPE is
the rotation angle. Using the summed hexatag embedding at position $p$, a
per-layer projection $W_{\delta}^{(\ell)}$ produces a per-frequency phase offset
\begin{equation}
\Delta\theta_{p,t} \;=\; \big(W_{\delta}^{(\ell)}(\textcolor{tagcolor}{\mathbf{E}^T_{\tau_p}} + \textcolor{tagcolor}{\mathbf{E}^N_{\nu_p}})\big)_t ,
\label{eq:modernbert_phase}
\end{equation}
which is added to the rotary angle \emph{before} rotation, for both the query at
position $m$ and the key at position $j$:
\begin{align}
\tilde{q}_{m,t} &\;=\; R\!\big(m\,\omega_t + \Delta\theta_{m,t}\big)\,q_{m,t}, \label{eq:modernbert_q}\\[2pt]
\tilde{k}_{j,t} &\;=\; R\!\big(j\,\omega_t + \Delta\theta_{j,t}\big)\,k_{j,t}. \label{eq:modernbert_k}
\end{align}
The effective angle between query $m$ and key $j$ at frequency $t$ is then
$(m-j)\,\omega_t + (\Delta\theta_{m,t}-\Delta\theta_{j,t})$, so the tag enters as
a syntactic phase shift on top of the positional one: the rotation that already
encodes relative distance is nudged by how the two positions' hexatags relate.
Since the tag enters only as a rotation angle, it leaves the query and key
magnitudes unchanged and perturbs only their direction (Appendix~\ref{app:modernbert}).

%% file: sections/results.tex
\section{Experimental Results}

\subsection{Experimental Settings and Datasets}
\label{sec:exp-settings}

Prior work injecting syntax into Transformers focused mainly on {\em intrinsic} evaluation---directly testing syntactic understanding \cite{Sartran_2022, xie-etal-2021-transformer, omote2019dependencyrelativepos, Yoshida_2022, zhao2024dependencytransformergrammarsintegrating, murty2023pushdownlayersencodingrecursive}. We argue these gains should also be assessed {\em extrinsically}, on downstream tasks where syntax is indirectly useful, and organize our experiments accordingly.

\paragraph{Intrinsic evaluation:} We test syntactic understanding on BLiMP \cite{warstadt2023blimpbenchmarklinguisticminimal} and SyntaxGym \cite{hu2020systematicassessmentsyntacticgeneralization}, both targeted minimal-pair benchmarks contrasting grammaticality (Table~\ref{tab:dataexamples}). BLiMP requires assigning higher probability to the grammatical sentence overall, whereas SyntaxGym compares probabilities only in the critical region where the ungrammaticality occurs.

Following prior work~\cite{zhao2024dependencytransformergrammarsintegrating, Sartran_2022},
we pretrain Transformer-XL on BLLIP-LG~\cite{charniak2000bllip} with their
hyperparameters, using the training splits of \citet{hu2020systematicassessmentsyntacticgeneralization}.\footnote{We do {\em not} use the dataset's syntactic annotations; we hexatag the sentences with an in-house parser (URL hidden for blind review).} We report BLiMP accuracy via sentence log-likelihood, SyntaxGym via the standard
suite-level voting protocol, and BLLIP-LG test perplexity to confirm
language-modeling ability is preserved. For our models, test sentences are
hexatagged once by the parser before scoring, so our perplexity is conditioned
on this single fixed parse, $p(x \mid \hat{T}(x))$, whereas joint syntactic LMs
report a marginal $p(x)$ approximated by summing over many candidate
parses.\footnote{This is the trade-off SiPE makes by design: a single parse at
inference instead of expensive marginalization (\autoref{fig:pareto}).}


In a second set of experiments, we pretrain three encoders: RoBERTa, DeBERTa, ModernBERT \cite{liu2019robertarobustlyoptimizedbert, he2023debertav3improvingdebertausing, warner2024smarterbetterfasterlonger} and Transformer-XL for 1M steps on a 50M-token, offline-hexatagged slice of WikiText-103 \cite{merity2016pointer}.\footnote{Our goal is to study syntactic embeddings on real downstream tasks, so we pretrain on WikiText and then fine-tune on GLUE. For encoders, we also re-use the same checkpoints for syntactic evaluation as a separate BLLIP-LG pretraining run with its own hyperparameter tuning was beyond our academic budget.} We evaluate these models on a hexatagged BLiMP \cite{warstadt2023blimpbenchmarklinguisticminimal}: using encoders, for each minimal pair we prefer the sentence with higher pseudo-log-likelihood (PLL) \cite{Salazar_2020} under the masked-LM objective; for the Transformer-XL decoder we instead use sentence log-likelihood (the sum of next-token log-probabilities). Accuracy is scored against BLiMP's gold labels.

\paragraph{Extrinsic evaluation:} 
To assess the downstream impact of injecting syntactic information directly in the transformer architecture, we carry out experiments on the GLUE benchmark~\cite{wang2019gluemultitaskbenchmarkanalysis}, which contains a suite of real-world NLP tasks. We finetune our three encoder models as well as Transformer-XL (which were all previously pre-trained with SiPE on Wikitext) on each task using standard GLUE hyperparameters for each model (refer to Tables \ref{tab:glue_base_encoders}, \ref{tab:glue_large_encoders} and \ref{tab:glue_decoder_txl} for results) and table \ref{tab:glue_hparams} in appendix \ref{sec:glue-hparams}. 


\subsection{Key Takeaways}
\label{takeaways}

\paragraph{Positional-pathway injection is the strongest recipe for a relative-PE decoder.} On Transformer-XL, injecting the SiPE prior into the positional pathway gives our best decoder result---SyntaxGym 80.60 ($+10.29$ over the vanilla token baseline) with perplexity cut from 18.63 to 16.95 (\autoref{tab:results_delta-dup1}). This carries through to GLUE, improving the TXL macro average from 68.17 to 73.78, a dramatic $+8.2\%$ relative gain (\autoref{tab:glue_decoder_txl}). Using only a single parse tree at inference, SiPE also surpasses parser-free approaches, TreeReg \cite{nandi2025sneakingsyntaxtransformerlanguage} and all Tree-Planted Transformer variants \cite{Yoshida_2024}, on both SyntaxGym and BLLIP-LG perplexity, cutting perplexity by roughly a quarter against the strongest such baseline (16.95 vs.\ 22.30 for TreeReg) and by more than half against the Tree-Planted Transformer variants (16.95 vs.\ 45.5--47.7) while matching or exceeding their syntactic generalization.

\paragraph{For encoders, the simpler input-embedding injection is better, with gains that strengthen out of distribution.} Across all three encoder families---absolute (RoBERTa), disentangled-relative (DeBERTa-v3), and rotary (ModernBERT), input-pathway SiPE consistently outperforms the base model on GLUE (\autoref{tab:glue_base_encoders}); the positional-pathway route also helps for the relative and rotary schemes but yields slightly smaller gains, so for encoders mixing the prior directly with input embeddings is sufficient and most reliable. All three encoders likewise improve on BLiMP---$+1.41\%$ (RoBERTa), $+2.27\%$ (DeBERTa-v3), $+2.32\%$ (ModernBERT) (\autoref{tab:blimp_comparison})---and these gains grow under continued pre-training on BLLIP-LG after WikiText, rising to $+1.87\%$, $+3.06\%$, and $+4.21\%$ respectively, indicating the prior generalizes beyond its original distribution rather than overfitting.
\footnote{The small BLiMP drops are expected: BLiMP scores the \emph{full}
sentence, so language-modeling ability dominates over the ungrammatical span
alone \cite{zhao2024dependencytransformergrammarsintegrating}. Most baselines in
\autoref{tab:results_delta-dup1} (PLM, Transformer Grammars, TreeReg, most TPT
variants) likewise fall below the vanilla token baseline on BLiMP.}

%% file: sections/layerwise.tex
\begin{figure}[t]
  \centering
  \includegraphics[width=\columnwidth]{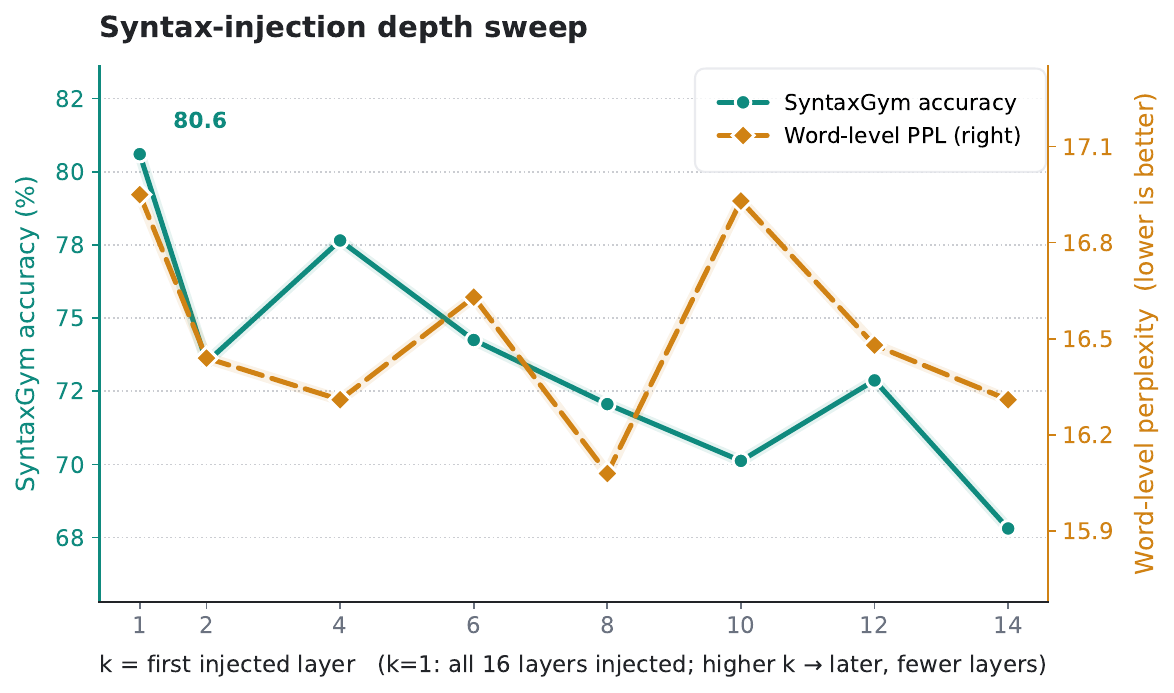}
  \caption{\footnotesize Layerwise SiPE injection sweep on Transformer-XL.
  Injecting from layer $k{=}1$ (all layers) is best; skipping the first
  layer already drops SyntaxGym accuracy sharply, and later entry points
  degrade syntactic generalization further.}
  \label{fig:layerwise_sweep}
\end{figure}

\section{Where Should Syntax Enter? A Layerwise Injection Study}
\label{sec:layerwise}

To understand at which depth syntax should interact with semantics, we
sweep the entry point of the positional pathway SiPE bias (PP-SiPE) in Transformer-XL. For a model with $N$
layers, we inject the bias from layer $k$ onwards: layers $1,\dots,k{-}1$
use the default positional information, while layers $k,\dots,N$ receive
the SiPE bias. We vary $k \in \{2, 4, 6, 8, 10, 12, 14\}$ and compare
injecting syntactic information in every layer (from $k{=}1$ onwards), measuring syntactic generalization
(SyntaxGym) and word-level perplexity on BLLIP\textsubscript{LG} (\autoref{fig:layerwise_sweep}).

Two findings emerge. First, syntax is most impactful at the very first
layer: full injection ($k{=}1$) is clearly best, and skipping just the
first layer ($k{=}2$) already causes a sharp drop in syntactic
generalization: SyntaxGym falls from $80.6$ to $73.5$. Second, and more
broadly, syntactic information matters most in the lower layers of a
decoder like Transformer-XL: injecting it in later layers
yields consistently weaker syntactic generalization. Together, these
results indicate that for autoregressive models the positional pathway
should carry syntactic structure from the earliest layer onward, rather
than being introduced in later layers.

Complementing this view of \emph{where} syntax should enter,
Appendix~\ref{app:attn-analysis} examines \emph{how} the injected prior
manifests in attention on BLiMP object--verb agreement: the encoders
redistribute verb$\rightarrow$object attention to mirror syntactic
adjacency, whereas Transformer-XL surfaces the prior only weakly in its
attention maps and instead converts it into the largest downstream
semantic (GLUE) gains of any model we train.

%% file: sections/conclusion.tex
\section{Conclusion}
We proposed a simple strategy to infuse syntactic information in any transformer architecture.
Our proposed modification is lightweight, adding only \(O(1,000)\) parameters per model via an additive prior embedding, yet it yields reliable gains in both intrinsic and extrinsic evaluations and multiple encoder and decoder transformer variants.

%% file: sections/appendix.tex
\appendix

\section{Preliminaries}
\label{sec:appendix}

\subsection{Subword-Level Tag Injection}
\label{subword-tag-injection}
Hexatag annotations are produced at the word level: every word $w_i$ in a
sentence carries a terminal tag $t_i$ and a nonterminal tag $n_i$ (drawn from
small vocabularies $\mathcal{T}$ and $\mathcal{N}$, respectively, where $|T| = 2$ and $|N| = 5$). The language model, however, operates on subword tokens: the tokenizer may split
$w_i$ into one or more subwords $s_{i,1}, \dots, s_{i,k_i}$. 
We must therefore specify how a word-level hexatag is associated with its corresponding subword positions.

We adopt a \emph{assign hexatag to first-subword only strategy}: the tag pair $(t_i, n_i)$ is attached to the position of the first subword $s_{i,1}$, and the remaining
subwords $s_{i,2}, \dots, s_{i,k_i}$ receive no tag (in our implementation this is equivalent to receiving a tag ID of $-100$). Concretely, for each
subword position $p$ in the flattened input sequence, we define the following
mask:
\begin{equation}
m_p =
\begin{cases}
1 & \text{if } p \text{ is the first subword} \\
  & \text{of some word } w_i, \\
0 & \text{otherwise.}
\end{cases}
\end{equation}
At subword positions with $m_p = 1$, the prior contributes a learned tag embedding to
augment the subword embedding; at positions with $m_p = 0$ (continuation subwords),
no prior is added, and the subword embedding is used as-is. Tag information
still reaches continuation subwords, but only indirectly, through
self-attention in subsequent layers.

This design has two practical benefits: (i) The total tag embedding signal
injected per word is invariant to its subword length $k_i$, avoiding
over-amplification for words that break into many subwords

(ii) it gives a clean one-to-one correspondence between words and
tagged subword positions, which is required by the auxiliary tag-prediction
objective: this objective is a per-position classification head that predicts
the (terminal, nonterminal) tag pair, and it only contributes loss at
positions where $m_p = 1$.

The \emph{assign hexatag to first-subword only} is also the simpler choice here. The natural alternative would be to assign the same hexatag to all $k_i$ subwords of a word, which would require supervising all pieces, contributing $k_i$ loss terms per word and thus over-weighting words that fragment into many subwords. We leave this variant to future work. 

The indicator $m_p$ thus controls both prior injection and supervision, giving each word exactly one tag signal at the input and one prediction at the output. The auxiliary objective takes one of two concrete forms in our experiments.
In autoregressive models, the head at position $p$ predicts the tags
$(\tau_{p+1}, \nu_{p+1})$ of the \emph{next} tagged position,
teacher-forced from gold left context. In masked language models, the head
at position $p$ predicts $(\tau_p, \nu_p)$ at masked positions only; wherever a
token is masked, its injected tag embedding is masked along with it (the $m_p$
term is zeroed), so the auxiliary head must infer the tag from surrounding
context rather than copy it from its own input.

The remainder of this section details the design choices and trade-offs for each
injection strategy we study. These fall into a few families: adding the
syntactic bias to the input embeddings; entangling it directly with the
positional encoding; keeping it disentangled and adding it as a separate term to
the attention score; injecting it as a separate attention-side bias; and
combining any of these. We examine these strategies across four model architectures that
span the major positional-encoding schemes: RoBERTa (encoder, absolute
positional embeddings), DeBERTa-v3 (encoder, relative positional embeddings),
ModernBERT (encoder, rotary positional embeddings), and Transformer-XL
(autoregressive decoder, relative positional embeddings).

\subsection{SiPE Position-Pathway Design Choices for Transformer-XL}
\label{app:txl_pebias_design}

The multiplicative position-pathway injection of section \ref{sec:txl_pebias} has two
design axes. \textbf{Layer sharing} controls whether a single projection
$\textcolor{tagcolor}{W_E}$ is reused across all layers or each layer learns its
own $\textcolor{tagcolor}{W_E^{(\ell)}}$. \textbf{Projection target} controls the dimension
$\textcolor{tagcolor}{W_E^{(\ell)}}$ projects the tag into before its inner
product with the query forms the alignment score $\textcolor{gatecolor}{c}$: either a
single attention head's $d_{\text{head}}$-dimensional subspace (a small map that
places the tag in the same space as that head's query, so $\textcolor{gatecolor}{c}$
is formed by the same query that produces $\textcolor{relposcolor}{\mathbf{BD}}$
and lands on the same per-head scale; the one projected vector is then shared
across all heads), or the full model dimension $d_{\text{model}}$ (a larger map
whose output is split into $n_{\text{head}}$ blocks, giving each head its own
projected tag). Table~\ref{tab:txl_pebias_ablation} reports all four
combinations.

\begin{table}[h]
\centering
\small
\renewcommand{\arraystretch}{1.2}
\begin{tabular}{@{}llcc@{}}
\toprule
\textbf{Sharing} & \textbf{Projection} & \textbf{BLiMP}\,$\uparrow$ & \textbf{SG}\,$\uparrow$ \\
\midrule
Shared     & $d_{\text{head}}$  & 73.95 & 78.80 \\
Per-layer  & $d_{\text{head}}$  & 74.15 & 75.84 \\
Shared     & $d_{\text{model}}$ & 73.95 & 75.09 \\
Per-layer  & $d_{\text{model}}$ & \textbf{74.01} & \textbf{80.60} \\
\bottomrule
\end{tabular}
\caption{Multiplicative position-pathway injection across the two design axes:
layer sharing (shared vs.\ per-layer $\textcolor{tagcolor}{W_E}$) and projection
target ($d_{\text{head}}$ head subspace vs.\ $d_{\text{model}}$ full dimension).
Per-layer projection into the full model dimension gives the best average across
BLiMP and SyntaxGym (SG); this is the configuration we adopt as SiPE (positional pathway) for Transformer-XL.}
\label{tab:txl_pebias_ablation}
\end{table}

For Transformer-XL, the best configuration we find is the per-layer projection into the full model dimension
(74.01 BLiMP, 80.60 SG). Neither axis dominates alone:
the gain comes from their pairing. Per-layer projections matter because the tag
correction is scaled by each layer's own $\textcolor{relposcolor}{\mathbf{BD}}$,
so a separate $\textcolor{tagcolor}{W_E^{(\ell)}}$ lets each layer specialize how
syntax modulates its positional preference. The full-dimension target matters
because it gives the tag a richer map than a single head's subspace allows.
Notably, the better projection target flips with the coupling: the multiplicative
coupling here works better with $d_{\text{model}}$, whereas the fully-disentangled coupling
of Appendix~\ref{app:txl_disentangled_design} works better in the $d_{\text{head}}$ head
subspace. The best target therefore depends on how the prior is coupled to the
position pathway, not on the projection axis in isolation.

Beyond these two axes, the form of the coupling itself matters: multiplicative coupling scales the position
term by the tag alignment (Eq.~\ref{eq:txl_sipe_mul}), while the disentangled
form adds the alignment as a standalone logit term
(section \ref{sssec:txl_disentangled}, Eq.~\ref{eq:txl_sipe_dis}). The multiplicative
variants are the strongest overall, but the ordering is informative:
disentangled injection already outperforms input-side injection
(Eq.~\ref{eq:txl_add}), and multiplicative coupling outperforms both. Entangling the
prior with position thus helps progressively more as the coupling becomes
gated by the query's existing offset alignment, so that syntax sharpens a
positional preference the model has already formed rather than adding an
unconditional bias.

\subsection{Fully-Disentangled Syntax Injection Design Choices for Transformer-XL}
\label{app:txl_disentangled_design}

The fully-disentangled variant of section \ref{sssec:txl_disentangled} adds the
tag--query alignment $\textcolor{gatecolor}{c}$ to the attention logit as a
standalone term (Eq.~\ref{eq:txl_sipe_dis}), leaving both the content term
$\mathbf{AC}$ and the position term $\textcolor{relposcolor}{\mathbf{BD}}$
exactly as in standard Transformer-XL. It is governed by the same two design
axes (layer sharing and projection target) as the multiplicative
position-pathway variant of Appendix~\ref{app:txl_pebias_design}, the only
difference being that there $\textcolor{gatecolor}{c}$ multiplies
$\textcolor{relposcolor}{\mathbf{BD}}$ whereas here it is added as a standalone
logit term:
$\textcolor{tagcolor}{W_E^{(\ell)}}$ may map the tag into the full model
dimension (then reshaped into per-head blocks) or directly into a single head's
$d_{\text{head}}$-dimensional subspace, and it may be shared across all layers or
learned per layer. Since neither axis has an obvious right answer a priori, we
ran all four combinations and report them in
Table~\ref{tab:txl_disentangled_ablation}.

\begin{table}[h]
\centering
\small
\renewcommand{\arraystretch}{1.2}
\begin{tabular}{@{}llcc@{}}
\toprule
\textbf{Sharing} & \textbf{Projection} & \textbf{BLiMP}\,$\uparrow$ & \textbf{SG}\,$\uparrow$ \\
\midrule
Shared     & $d_{\text{head}}$  & 74.17 & 77.72 \\
Shared     & $d_{\text{model}}$ & 75.03 & 76.10 \\
Per-layer  & $d_{\text{model}}$ & 73.51 & 77.20 \\
Per-layer  & $d_{\text{head}}$  & \textbf{74.72} & \textbf{78.72} \\
\bottomrule
\end{tabular}
\caption{Fully-disentangled injection across the two projection axes: layer
sharing (a single $\textcolor{tagcolor}{W_E}$ shared across all layers vs.\ a
per-layer $\textcolor{tagcolor}{W_E^{(\ell)}}$) and projection target (mapping
the tag into a single head's $d_{\text{head}}$-dimensional subspace vs.\ the
full model dimension $d_{\text{model}}$, reshaped into per-head blocks).
Per-layer projection into the head subspace gives the best average across BLiMP
and SyntaxGym (SG); this is the configuration reported in the main results.}
\label{tab:txl_disentangled_ablation}
\end{table}

The best disentangled configuration is per-layer projection into the head
subspace (74.72 BLiMP, 78.72 SG; Table~\ref{tab:txl_disentangled_ablation}). In
terms of syntactic generalization performance, it outperforms input-side
injection (73.82 BLiMP, 76.97 SG) but lags behind the multiplicative coupling
(\S\ref{sssec:txl_entangled}). We thus find that the least to most performant
design choices for entangling the syntactic prior are: input-side $<$
fully disentangled $<$ multiplicative.

\subsection{Attention-Side and Combined Injection Design Choices for Transformer-XL}
\label{app:txl_attn_design}

\paragraph{Relation to the Shaw formulation:}
The key and value biases of Eqs.~\ref{eq:txl_attn_k}--\ref{eq:txl_attn_v} adapt ~\citet{shaw2018selfattentionrelativepositionrepresentations}, who add
learned per-pair vectors $\mathbf{a}^K_{ij}$ and $\mathbf{a}^V_{ij}$ to the
key-side logit and the value-side aggregation,
$e_{ij}\mathrel{+}=\mathbf{q}_i^{\top}\mathbf{a}^K_{ij}$ and
$\mathbf{z}_i\mathrel{+}=\sum_j\alpha_{ij}\mathbf{a}^V_{ij}$. We change two
things. First, our bias is indexed by the key position $j$ alone rather than by
the relative pair $(i,j)$, so the tag at $j$ contributes a single per-key vector
that every query sees; this is what keeps the added cost linear in sequence
length rather than quadratic. Second, the bias is conditioned on hexatag
identity rather than relative offset, and is learned jointly with the language
model. The terminal and
nonterminal tag embeddings are projected separately and summed, and a
position-validity mask $m_j$ zeroes the contribution wherever the tag label is
$\texttt{-100}$. The tag bias is formed from the unmodified query $\mathbf{q}^{(\ell)}_i$, not
$\mathbf{q}^{(\ell)}_i+\mathbf{u}_n$ as in the content term, so it bypasses the
global content bias $\mathbf{u}_n$ of
\citet{dai2019transformerxlattentivelanguagemodels}. And since we set
$\texttt{mem\_len}=0$, it spans the same keys as intra-segment attention, so no
padding is needed for memory positions.

This places our design between the dependency-distance bias of Omote et
al.~\cite{omote2019dependencyrelativepos} and the root-to-node path embedding of
Xie et al.~\cite{xie-etal-2021-transformer}. Both put syntactic
structure inside attention rather than in the residual stream, but our bias is
keyed on hexatag identity rather than a tree-distance proxy, is a single per-key
term rather than a pairwise one, and is applied in a causally masked
unidirectional language model rather than a bidirectional encoder.

\paragraph{Sharing across layers.}
The one design consideration here is whether $\mathbf{W}^{\text{tag}}_K$ and
$\mathbf{W}^{\text{tag}}_V$ are shared across all layers or learned per layer.
In the shared form a single pair is reused at every layer; in the per-layer
form each of the $L$ layers owns its own pair, giving $L$ independent key and
value maps and letting each layer specialize its tag-to-attention routing, in
the same way the content projections $\mathbf{W}_Q,\mathbf{W}_K,\mathbf{W}_V$
already specialize per layer. The per-layer form is the stronger of the two on
syntactic evaluations, both for the attention-side injection on its own and for
its combination with the input-side injection of Eq.~\ref{eq:txl_add}, so we use
it wherever the attention-side biases appear.


\subsection{DeBERTa-v3: Disentangled Relative Positional Embeddings and Tag Injection}
\label{app:deberta_injection}

DeBERTa-v3~\cite{he2023debertav3improvingdebertausing} does not use
absolute positional embeddings at the input level. Instead it delivers
positional information through \emph{disentangled relative attention}: each pair
of positions $(i,j)$ contributes three distinct terms to the attention score,
only one of which depends on token content alone. This raises a natural question
for our setting: where do tag priors enter when there is no input-level position
embedding to sit alongside?

\paragraph{Background: disentangled attention.}
For a query at position $i$ and a key at position $j$, the score decomposes into
a content-to-content (c2c), a content-to-position (c2p), and a
position-to-content (p2c) term,
\begin{align}
A_{ij} \;=\;& \underbrace{\mathbf{q}_i^{\top} \mathbf{k}_j}_{\text{c2c}}
   \;+\; \underbrace{\mathbf{q}_i^{\top} \mathbf{W}_{c2p} \,\textcolor{relposcolor}{\mathbf{r}_{\delta(i,j)}}}_{\text{c2p}} \notag \\
   &\;+\; \underbrace{\mathbf{k}_j^{\top} \mathbf{W}_{p2c} \,\textcolor{relposcolor}{\mathbf{r}_{\delta(j,i)}}}_{\text{p2c}},
\label{eq:deberta_attn_app}
\end{align}
where $\mathbf{q}_i = \mathbf{W}_Q \mathbf{h}^{(0)}_i$ and
$\mathbf{k}_j = \mathbf{W}_K \mathbf{h}^{(0)}_j$. The
\textcolor{relposcolor}{relative-position embeddings $\mathbf{r}_{\delta(\cdot,\cdot)}$}
are pulled from a learned table indexed by bucketed signed offsets, and
$\mathbf{W}_{c2p}, \mathbf{W}_{p2c}$ project content vectors into this
relative-position subspace.\footnote{The buckets $\delta$ collapse pairs of
positions with similar offsets into shared rows of the relative-position table,
following \cite{shaw2018selfattentionrelativepositionrepresentations}: small
offsets $\{-3,-2,-1,0,1,2,3\}$ each receive their own bucket, while larger
offsets are progressively merged --- e.g.\ $\{+9,\ldots,+15\}$ might collapse
into a single bucket --- so the table stays small while preserving fine-grained
discrimination near the diagonal.}

\paragraph{Input-side injection: where priors enter, and where they do not.}
Under input-side injection (Eq.~\ref{eq:deberta_input}) each input vector
carries the summed tag prior, so the queries and keys carry it into the score
through the content stream:
\begin{align}
\mathbf{q}_i \;=\;& \mathbf{W}_Q \!\Bigl(\mathbf{e}^{\text{tok}}_i + \mathbf{e}^{\text{seg}}_i
   + m_i \!\cdot\! \bigl(\textcolor{tagcolor}{\mathbf{E}^{T}_{\tau_i}} + \textcolor{tagcolor}{\mathbf{E}^{N}_{\nu_i}}\bigr)\!\Bigr), \label{eq:deberta_q_app} \\
\mathbf{k}_j \;=\;& \mathbf{W}_K \!\Bigl(\mathbf{e}^{\text{tok}}_j + \mathbf{e}^{\text{seg}}_j
   + m_j \!\cdot\! \bigl(\textcolor{tagcolor}{\mathbf{E}^{T}_{\tau_j}} + \textcolor{tagcolor}{\mathbf{E}^{N}_{\nu_j}}\bigr)\!\Bigr). \label{eq:deberta_k_app}
\end{align}
Reading Eqs.~\ref{eq:deberta_attn_app}--\ref{eq:deberta_k_app} together makes the
design visible. The \textcolor{tagcolor}{tag priors} sit inside
$\mathbf{h}^{(0)}_i$ and $\mathbf{h}^{(0)}_j$ and therefore propagate through
$\mathbf{W}_Q,\mathbf{W}_K$ into all three score terms via the content
projections. The \textcolor{relposcolor}{relative-position table}, by contrast,
appears only in c2p and p2c, and is purely a function of the offset. The two
streams enter through entirely separate pathways and meet only when their score
terms are summed to form the attention logit $A_{ij}$; the relative-position components themselves --- the
bucketing, the $\textcolor{relposcolor}{\mathbf{r}_{\delta}}$ table, and the
$\mathbf{W}_{c2p},\mathbf{W}_{p2c}$ projections --- are left entirely unmodified.

\textbf{Tag priors interact with relative position through the content stream
alone}, which is what makes the design portable across positional schemes. The
tag tables $\textcolor{tagcolor}{\mathbf{E}^{T}},\textcolor{tagcolor}{\mathbf{E}^{N}}$
are constructed similarly to RoBERTa: uniformly initialized, masked by the
first-subword indicator $m_p$, and added inside the input layer norm.

\paragraph{Position-pathway injection:}
The position-pathway variant of section \ref{sssec:deberta_pebias} leaves the
input clean and instead multiplies the two relative-position terms (c2p and p2c,
grouped as $S^{\text{pos}}$ in Eq.~\ref{eq:deberta_score}) by the tag--query
alignment $c_{ij}$ of Eq.~\ref{eq:deberta_c}, leaving the content-to-content term
untouched. Here $c_{ij}$ is formed from the content query $Q^c_i$ and the
per-layer projection $W_{E}^{(\ell)}$ of the summed hexatag embedding, and the
relative-position table and the $\mathbf{W}_{c2p},\mathbf{W}_{p2c}$ projections
are left unmodified. This is the direct DeBERTa analogue of the multiplicative
Transformer-XL injection: the prior scales the position pathway and leaves the
content pathway untouched with the only difference being that DeBERTa's position
pathway carries two terms (c2p and p2c) rather than Transformer-XL's single
$\mathbf{BD}$.

\subsection{ModernBERT: Rotary Injection Details}
\label{app:modernbert}

\paragraph{Portability of input-side injection:}
The unifying observation across the three encoders is that input-level
injection commutes with whatever positional mechanism the model already
uses, whether absolute (RoBERTa), relative (DeBERTa), or rotary (ModernBERT).

\paragraph{Identity at initialization.}
With $W_{\delta}^{(\ell)}\!\to\!\mathbf{0}$ we have $\Delta\theta\!\to\!0$ in
Eq.~\ref{eq:modernbert_phase}, so
Eqs.~\ref{eq:modernbert_q}--\ref{eq:modernbert_k} reduce to plain RoPE:
training starts from an unmodified ModernBERT, exactly as in the
Transformer-XL and DeBERTa variants.


\section{Time Complexity}
\label{sec:time-complexity}
SiPE adds two small embedding tables for terminal and non-terminal tags, $\mathbf{E}^{T} \in \mathbb{R}^{2 \times d}$ and $\mathbf{E}^{N} \in \mathbb{R}^{5 \times d}$, totaling $7d$ parameters per pathway ($\sim$5K at $d=768$, $\sim$7K at $d=1024$), well under $0.01\%$ of the LM's total parameters\footnote{For comparison, a single attention layer in RoBERTa-base contains $\sim$2.4M parameters; the priors add less than $0.3\%$ of a single layer's parameters.}.
Looking up and adding these embeddings costs $\mathcal{O}(Ld)$ per sequence, which is lower-order than self-attention's $\mathcal{O}(L^{2}d)$ cost. Position-pathway
variants introduce tag--query interactions with $\mathcal{O}(L^{2}d)$ complexity, matching the asymptotic cost of self-attention. SiPE therefore preserves the underlying Transformer's asymptotic complexity. At inference, a single auxiliary parser pass supplies the hexatags, while two lightweight token-level heads add only linear overhead beyond the parser encoder.



\section{Analyzing Attention Patterns on Object--Verb Agreement Tasks}
\label{app:attn-analysis}

A natural desideratum for SiPE-pretrained models is that they should up-weight
attention between tokens that are linearly distant but syntactically adjacent
(distance $1$ in the dependency tree). We probe this on the BLiMP
\textit{Causative} split for three model architectures spanning the three
positional encoding families: \texttt{RoBERTa-base} (absolute),
\texttt{ModernBERT-base} (rotary), and \texttt{Transformer-XL} (relative), each
against its SiPE counterpart. Verb$\rightarrow$object attention is averaged
across all heads; the layers we average over differ by architecture. For
Transformer-XL, our layerwise injection study (section\ref{sec:layerwise}) shows
that syntax matters at \emph{every} layer, so we average over all of its
layers. We have not run the corresponding sweep for the encoders, so we
instead follow \citet{tenney-etal-2019-bert}, who find that ``BERT's
intermediate layers encode a rich hierarchy of linguistic information, with
surface features at the bottom, syntactic features in the middle and semantic
features at the top,'' and therefore average over each encoder's middle layer
band.\footnote{Layers 5--8 for the 12-layer \texttt{RoBERTa-base} and layers
9--12 for the 22-layer \texttt{ModernBERT-base}; all layers 1--16 for the
\texttt{Transformer-XL} we pre-train.}

\begin{figure}[t]
    \centering
    \includegraphics[width=\columnwidth]{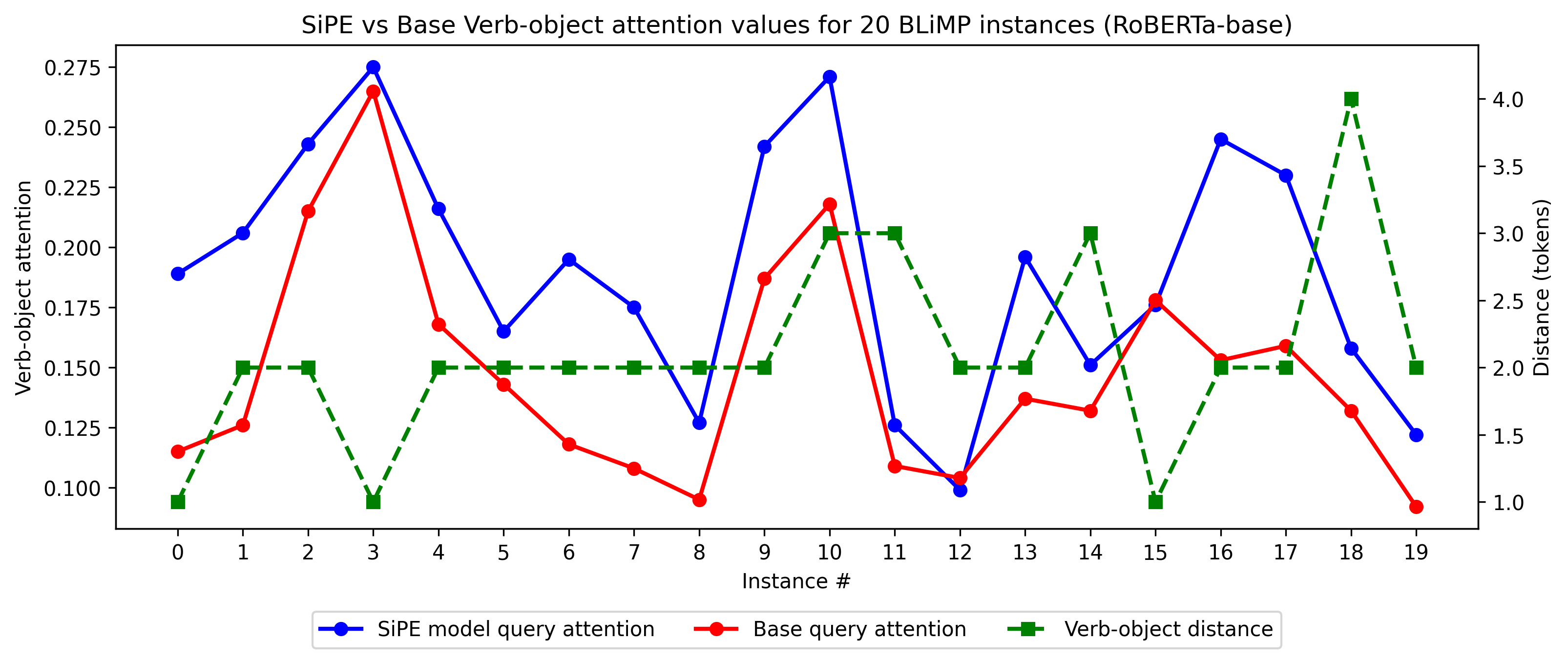}
    \caption{\footnotesize{Verb$\rightarrow$object attention on BLiMP \cite{warstadt2023blimpbenchmarklinguisticminimal} for \texttt{RoBERTa-base} vs.\ \texttt{RoBERTa-base+SiPE} (input pathway injection), both pre-trained on the 50M-token WikiText slice (\autoref{sec:exp-settings}). On the subset of \textit{Causative} examples where SiPE assigns higher PLL to the grammatical sentence but the base model does not, verb$\rightarrow$object attention is higher under SiPE in 90\% of cases (Appendix~\ref{app:attn-analysis}).}}
    \label{fig:syntactic_vs_base_attention}
\end{figure}

\begin{figure}[t]
    \centering
    \includegraphics[width=\columnwidth]{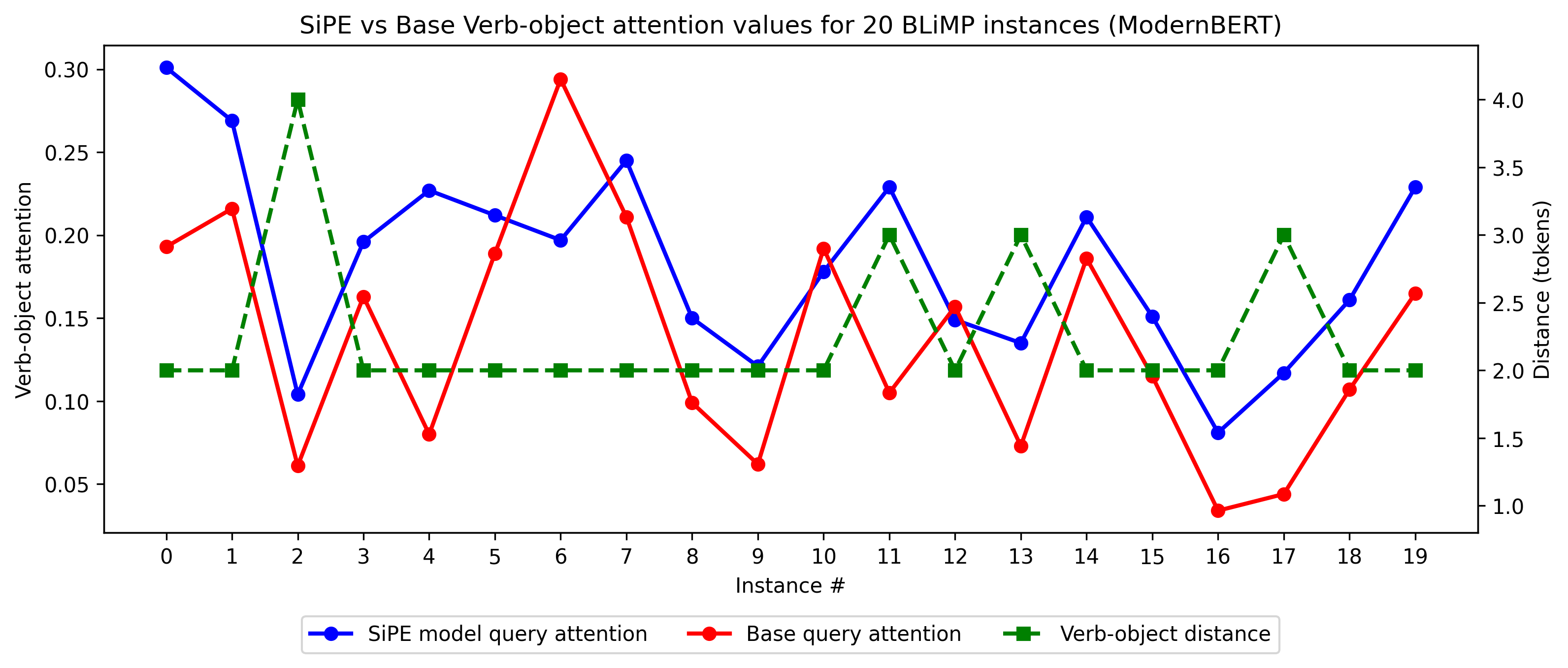}
    \caption{\footnotesize{Verb$\rightarrow$object attention on BLiMP for \texttt{ModernBERT-base} vs.\ \texttt{ModernBERT-base+SiPE} (input pathway injection), both pre-trained on the 50M-token WikiText slice (\autoref{sec:exp-settings}). On \textit{Causative} examples where SiPE is correct and the base model is not, the SiPE model assigns higher verb$\rightarrow$object attention in 17/20 (85\%) cases, mirroring the RoBERTa pattern despite ModernBERT's rotary positional encoding.}}
    \label{fig:modernbert_syntactic_vs_base_attention}
\end{figure}

\begin{figure}[t]
    \centering
    \includegraphics[width=\columnwidth]{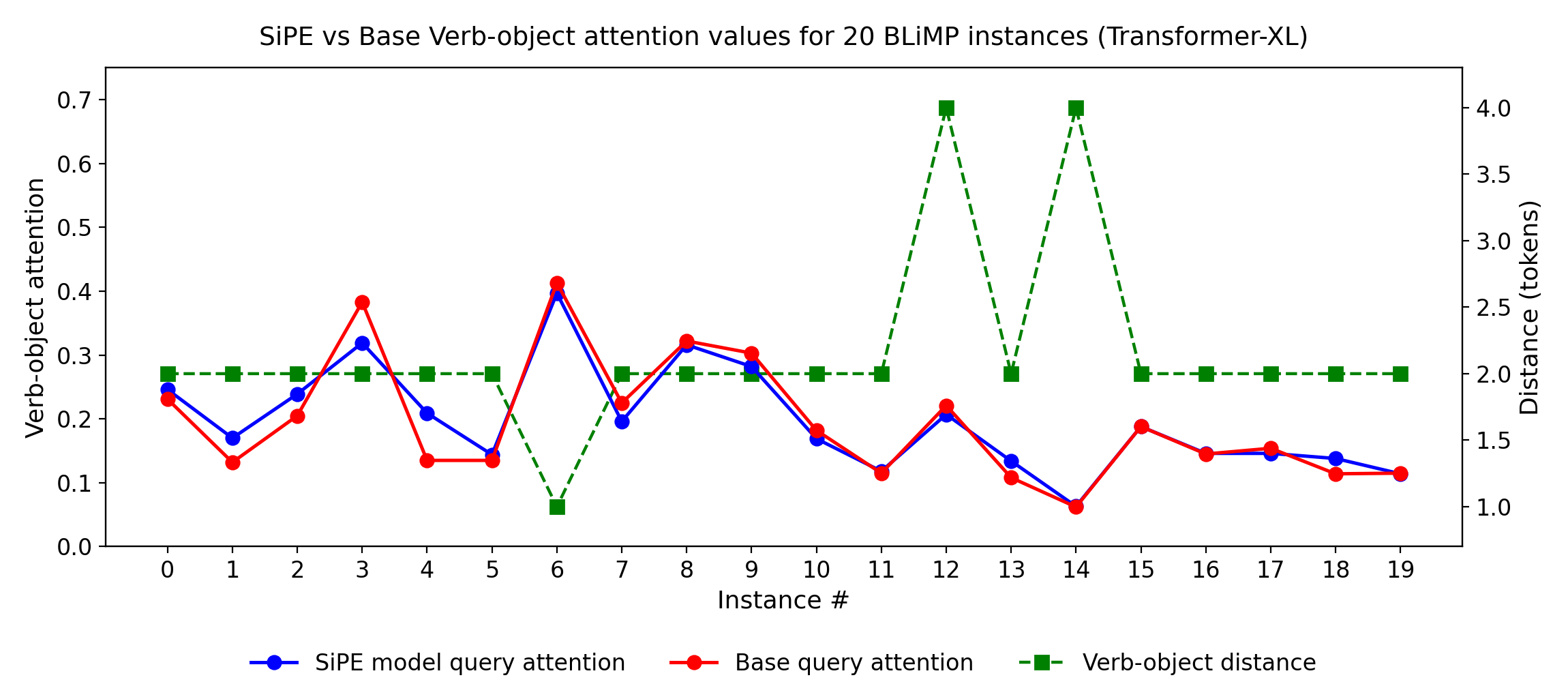}
    \caption{\footnotesize{Verb$\rightarrow$object attention on BLiMP for \texttt{Transformer-XL} vs.\ \texttt{Transformer-XL+SiPE} (positional pathway injection), both pre-trained on the 50M-token WikiText slice (\autoref{sec:exp-settings}), averaged over all $16$ layers. On \textit{Causative} examples where SiPE assigns higher sentence log-likelihood to the grammatical sentence but the base model does not, SiPE places higher verb$\rightarrow$object attention in $11/20$ ($55\%$) cases: the same direction as the encoders (Figs.~\ref{fig:syntactic_vs_base_attention} and~\ref{fig:modernbert_syntactic_vs_base_attention}), but a much weaker majority. The green dashed line (right axis) marks the verb$\rightarrow$object surface distance.}}
    \label{fig:txl_syntactic_vs_base_attention}
\end{figure}

Across all three families, SiPE attains higher overall BLiMP accuracy than the
corresponding base model. To understand how this syntactic information influences attention, we restrict the
analysis to the subset of minimal pairs where the SiPE model is correct and the base model is not. We manually inspect 20 examples from this subset. We find that \texttt{RoBERTa+SiPE} places higher
verb$\rightarrow$object attention than its base model in $90\%$ of cases
(Figure~\ref{fig:syntactic_vs_base_attention}) and \texttt{ModernBERT+SiPE}
does so in $85\%$ of cases (Figure~\ref{fig:modernbert_syntactic_vs_base_attention}).
\texttt{Transformer-XL} moves in the same direction but far more weakly: only a bare
majority ($11/20$ ($55\%$)) of such cases shift attention toward the object
(Figure~\ref{fig:txl_syntactic_vs_base_attention}), where the encoders push it
almost every time. Moreover, in over $80\%$ of the subset of 20 examples we examine, the object is nonadjacent
to the verb, indicating a redirection of attention weights to the syntactically relevant object, even when it is linearly distant from the verb. This suggests that SiPE effectively leverages syntactic information to guide attention in a way that captures long-range dependencies, which is crucial for understanding complex sentence structures.

The weak Transformer-XL effect poses an open question. The encoders appear to
spend most of the syntactic prior directly on attention, redistributing mass to mirror
syntactic adjacency. Transformer-XL barely does so in its attention maps, yet it
converts the same signal into the largest downstream gain of any model we train
($+8.2\%$ relative on GLUE; Table~\ref{tab:glue_decoder_txl}). This suggests the
decoder routes more of the enhanced syntactic knowledge beyond attention, towards semantic
processing.



\begin{figure*}[t]
    \centering
    \includegraphics[width=0.88\textwidth]{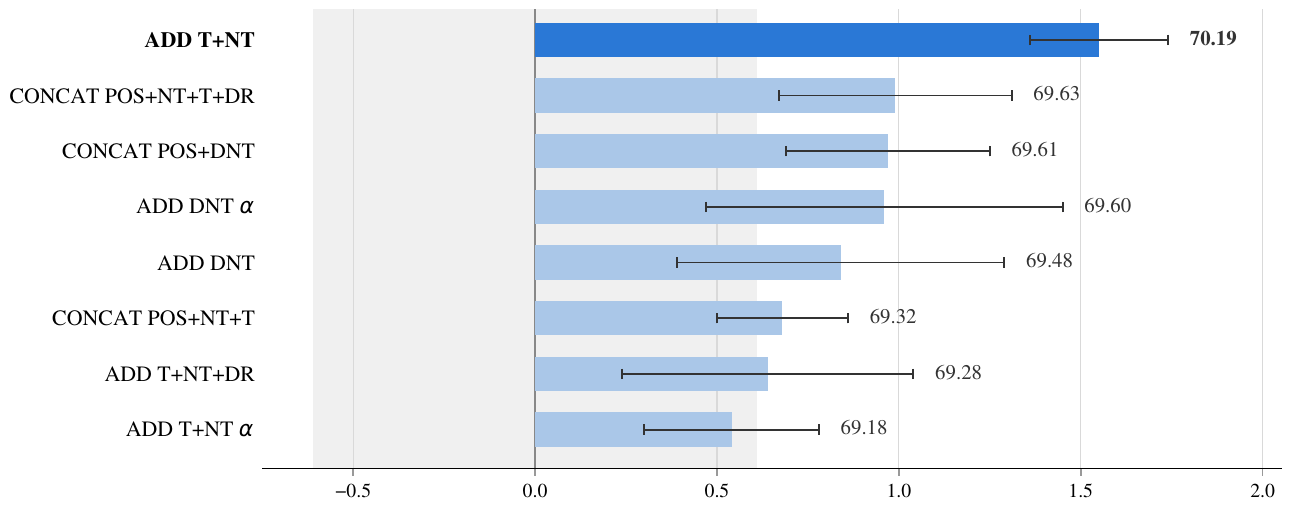}
    \caption{Average GLUE performance across prior-injection variants for RoBERTa-base, shown as the improvement over the no-prior baseline ($68.64 \pm 0.61$, shaded band at zero); bar-end labels give absolute scores and whiskers denote one standard deviation over 3 seeds. The strongest downstream performance is achieved by directly adding terminal and non-terminal prior embeddings to the absolute positional embedding model (\textbf{ADD T+NT}), outperforming concatenation, weighted addition, and variants using full dependency labels (the residual-injection variant is evaluated separately in Table~\ref{tab:other_combinations}). In these experiments, we set $\alpha=0.5$.}
    \label{fig:glue_avg_ablation}
\end{figure*}

\begin{figure*}[t]
    \centering
    \includegraphics[width=0.95\textwidth]{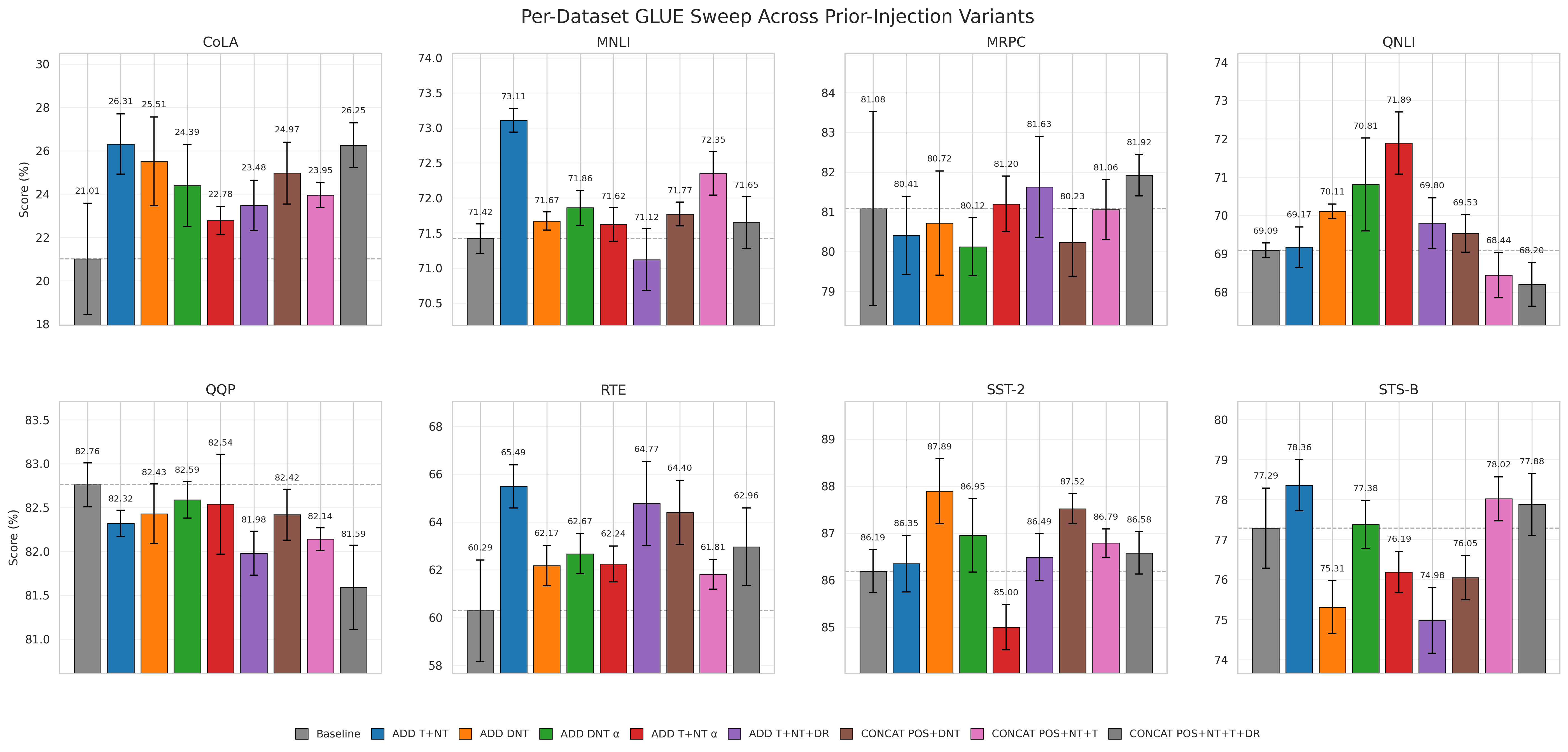}
    \caption{GLUE performance averaged across 3 random seeds \textbf{per dataset} under various prior-injection variants for RoBERTa-base. The strongest downstream performance is achieved by directly adding terminal and non-terminal prior embeddings to the absolute positional embedding model, outperforming concatenation, weighted addition, and variants using full dependency labels (the residual-injection variant is evaluated separately in Table~\ref{tab:other_combinations}).}
    \label{fig:deprel_per_dataset}
\end{figure*}

\begin{table*}[t]
\centering
\small
\setlength{\tabcolsep}{4pt}
\renewcommand{\arraystretch}{1.1}
\begin{tabular}{lcccc}
\toprule
\textbf{Task} & \textbf{Baseline} & \textbf{ADD\_T\_NT} & \textbf{CONCAT\_T\_NT} & \textbf{ADD\_T\_NT (Residual Connection)} \\
\midrule
CoLA  & 31.41\,{\scriptsize(±2.02)} & 31.77\,{\scriptsize(±0.71)} \textcolor{green}{$\uparrow$} & 26.85\,{\scriptsize(±1.06)} \textcolor{red}{$\downarrow$} & 32.35\,{\scriptsize(±0.79)} \textcolor{green}{$\uparrow$} \\
SST-2 & 87.23\,{\scriptsize(±0.79)} & 87.22\,{\scriptsize(±0.52)} \textcolor{red}{$\downarrow$}  & 87.27\,{\scriptsize(±0.35)} \textcolor{green}{$\uparrow$} & 87.42\,{\scriptsize(±0.19)} \textcolor{green}{$\uparrow$} \\
QQP   & 83.24\,{\scriptsize(±0.21)} & 83.36\,{\scriptsize(±0.11)} \textcolor{green}{$\uparrow$} & 82.60\,{\scriptsize(±0.22)} \textcolor{red}{$\downarrow$}  & 83.13\,{\scriptsize(±0.15)} \textcolor{red}{$\downarrow$} \\
QNLI  & 77.65\,{\scriptsize(±0.50)} & 78.24\,{\scriptsize(±0.27)} \textcolor{green}{$\uparrow$} & 70.79\,{\scriptsize(±0.70)} \textcolor{red}{$\downarrow$}  & 75.47\,{\scriptsize(±0.54)} \textcolor{red}{$\downarrow$} \\
MNLI  & 73.03\,{\scriptsize(±0.33)} & 74.05\,{\scriptsize(±0.17)} \textcolor{green}{$\uparrow$} & 72.89\,{\scriptsize(±0.31)} \textcolor{red}{$\downarrow$}  & 73.90\,{\scriptsize(±0.14)} \textcolor{green}{$\uparrow$} \\
RTE   & 65.76\,{\scriptsize(±0.74)} & 64.66\,{\scriptsize(±0.56)} \textcolor{red}{$\downarrow$}  & 64.74\,{\scriptsize(±1.45)} \textcolor{red}{$\downarrow$}  & 65.46\,{\scriptsize(±0.45)} \textcolor{red}{$\downarrow$} \\
STSB  & 79.11\,{\scriptsize(±0.30)} & 81.41\,{\scriptsize(±0.13)} \textcolor{green}{$\uparrow$} & 79.36\,{\scriptsize(±0.26)} \textcolor{green}{$\uparrow$} & 80.41\,{\scriptsize(±0.47)} \textcolor{green}{$\uparrow$} \\
MRPC  & 80.96\,{\scriptsize(±0.44)} & 82.02\,{\scriptsize(±0.85)} \textcolor{green}{$\uparrow$} & 82.76\,{\scriptsize(±0.72)} \textcolor{green}{$\uparrow$} & 81.06\,{\scriptsize(±0.96)} \textcolor{green}{$\uparrow$} \\
\midrule
\textbf{Average} & \textbf{72.30} & \textbf{72.84} \textcolor{green}{$\uparrow$} & \textbf{70.91} \textcolor{red}{$\downarrow$} & \textbf{72.40} \textcolor{green}{$\uparrow$} \\
\bottomrule
\end{tabular}
\caption{RoBERTa performance on GLUE under alternative strategies for combining syntactic priors. \textbf{ADD\_T\_NT} adds the terminal prior $\mathbf{t}_i$ and non-terminal prior $\mathbf{n}_i$ directly to the token representation (i.e., additive fusion at the embedding layer). \textbf{CONCAT\_T\_NT} concatenates the original embedding (including standard positional information) with $\mathbf{t}_i$ and $\mathbf{n}_i$, then applies a learned linear down-projection to the model dimension before feeding the encoder. \textbf{ADD\_T\_NT (Residual connection)} injects $(\mathbf{t}_i+\mathbf{n}_i)$ into the residual stream before the multi-head self-attention sublayer. Results are averaged over 3 random seeds. Green arrows indicate improvements over the baseline and red arrows indicate degradations. Overall, \textbf{ADD\_T\_NT} yields the strongest macro-average (72.84\%), while concatenation is the weakest configuration (70.91\%), with particularly large drops on QNLI and CoLA.}
\label{tab:other_combinations}
\end{table*}

\begin{table*}[t]
\centering
\small
\setlength{\tabcolsep}{4pt}
\renewcommand{\arraystretch}{1.1}
\begin{tabular}{lcccc}
\toprule
\textbf{Task} & \textbf{Baseline} & \textbf{$\alpha$ = 0.1} & \textbf{$\alpha$ = 0.5} & \textbf{$\alpha$ = 0.9} \\
\midrule
CoLA  & 31.41\,{\scriptsize(±2.02)} & 29.40\,{\scriptsize(±1.62)} \textcolor{red}{$\downarrow$} & 29.89\,{\scriptsize(±0.71)} \textcolor{red}{$\downarrow$} & 30.37\,{\scriptsize(±0.54)} \textcolor{red}{$\downarrow$} \\
SST-2 & 87.23\,{\scriptsize(±0.79)} & 86.93\,{\scriptsize(±0.34)} \textcolor{red}{$\downarrow$} & 86.81\,{\scriptsize(±0.29)} \textcolor{red}{$\downarrow$} & 88.13\,{\scriptsize(±0.06)} \textcolor{green}{$\uparrow$} \\
QQP   & 83.24\,{\scriptsize(±0.21)} & 83.04\,{\scriptsize(±0.08)} \textcolor{red}{$\downarrow$} & 83.01\,{\scriptsize(±0.11)} \textcolor{red}{$\downarrow$} & 83.66\,{\scriptsize(±0.48)} \textcolor{green}{$\uparrow$} \\
QNLI  & 77.65\,{\scriptsize(±0.50)} & 77.08\,{\scriptsize(±0.73)} \textcolor{red}{$\downarrow$} & 73.72\,{\scriptsize(±0.22)} \textcolor{red}{$\downarrow$} & 76.45\,{\scriptsize(±2.00)} \textcolor{red}{$\downarrow$} \\
MNLI  & 73.03\,{\scriptsize(±0.33)} & 72.58\,{\scriptsize(±0.02)} \textcolor{red}{$\downarrow$} & 73.50\,{\scriptsize(±0.37)} \textcolor{green}{$\uparrow$} & 73.07\,{\scriptsize(±0.05)} \textcolor{green}{$\uparrow$} \\
RTE   & 65.76\,{\scriptsize(±0.74)} & 66.25\,{\scriptsize(±0.18)} \textcolor{green}{$\uparrow$} & 64.98\,{\scriptsize(±0.62)} \textcolor{red}{$\downarrow$} & 66.07\,{\scriptsize(±1.44)} \textcolor{green}{$\uparrow$} \\
STSB  & 79.11\,{\scriptsize(±0.30)} & 78.38\,{\scriptsize(±0.03)} \textcolor{red}{$\downarrow$} & 79.64\,{\scriptsize(±0.51)} \textcolor{green}{$\uparrow$} & 78.25\,{\scriptsize(±0.14)} \textcolor{red}{$\downarrow$} \\
MRPC  & 80.96\,{\scriptsize(±0.44)} & 82.77\,{\scriptsize(±0.09)} \textcolor{green}{$\uparrow$} & 81.28\,{\scriptsize(±1.51)} \textcolor{green}{$\uparrow$} & 80.42\,{\scriptsize(±0.04)} \textcolor{red}{$\downarrow$} \\
\midrule
\textbf{Average} & \textbf{72.30} & \textbf{72.05} \textcolor{red}{$\downarrow$} & \textbf{71.60} \textcolor{red}{$\downarrow$} & \textbf{72.05} \textcolor{red}{$\downarrow$} \\
\bottomrule
\end{tabular}
\caption{Effect of initializing the interpolation gate $\alpha$ for RoBERTa under the \textbf{ADD\_T\_NT} setting. We initialize $\alpha \in \{0.1, 0.5, 0.9\}$ and allow it to be updated during pretraining via backpropagation; all results report downstream GLUE performance after pretraining. Despite task-level variability (notably on QNLI), none of the $\alpha$ initializations improves upon the baseline on average, and all are weaker than the simple additive fusion.}
\label{tab:alpha_combinations}
\end{table*}

\section{Hyperparameters}
\label{sec:glue-hparams}

Table~\ref{tab:glue_hparams} reports the task-specific fine-tuning hyperparameters used for each backbone. Unless noted otherwise, all models use AdamW with $(\beta_1,\beta_2)=(0.9,0.999)$ and $\epsilon=10^{-6}$.

\begin{table}[H]
\centering
\scriptsize
\setlength{\tabcolsep}{5pt}
\renewcommand{\arraystretch}{1.0}
\begin{tabular}{@{}lcc@{}}
\toprule
& \textbf{LR} & \textbf{WD} \\
\midrule
\rowcolor{gray!15}
\multicolumn{3}{@{}l}{\textit{RoBERTa-base/large \& DeBERTa-v3-base/large}} \\
\quad all tasks & $1\mathrm{e}{-5}$ & $0.1$ \\
\rowcolor{gray!15}
\multicolumn{3}{@{}l}{\textit{ModernBERT-base/large (per task)}} \\
\quad CoLA  & $8\mathrm{e}{-5}$ & $1\mathrm{e}{-6}$ \\
\quad SST-2 & $8\mathrm{e}{-5}$ & $1\mathrm{e}{-5}$ \\
\quad MRPC  & $5\mathrm{e}{-5}$ & $5\mathrm{e}{-6}$ \\
\quad STS-B & $8\mathrm{e}{-5}$ & $5\mathrm{e}{-6}$ \\
\quad QQP   & $5\mathrm{e}{-5}$ & $5\mathrm{e}{-6}$ \\
\quad MNLI  & $5\mathrm{e}{-5}$ & $5\mathrm{e}{-6}$ \\
\quad QNLI  & $8\mathrm{e}{-5}$ & $5\mathrm{e}{-6}$ \\
\quad RTE   & $5\mathrm{e}{-5}$ & $1\mathrm{e}{-5}$ \\
\bottomrule
\end{tabular}
\caption{GLUE fine-tuning hyperparameters. Only ModernBERT's LR/WD vary by task;
RoBERTa and DeBERTa-v3 use LR $1\mathrm{e}{-5}$, WD $0.1$ throughout. All runs:
$20$ epochs, AdamW, $\epsilon=10^{-6}$.}
\label{tab:glue_hparams}
\end{table}

\section{Alternative Embedding Combination Strategies}
\label{app:alternate-combination-appendix}

Throughout the paper, each token carries just two hexatag priors: a terminal tag
$\mathbf{t}_i = \mathbf{E}^T_{\tau_i}$ and a non-terminal tag
$\mathbf{n}_i = \mathbf{E}^N_{\nu_i}$, read from small learned tables ($2\times D$
for the terminals, $5\times D$ for the non-terminals). But is this the richest
syntactic signal we could inject? A natural alternative is the token's full
\emph{dependency-relation} label (\textsc{deprel}; 40 labels such as
\texttt{nsubj} and \texttt{dobj}), which would come from a much larger
$40\times D$ table, giving a per-token embedding $\mathbf{deprel}_i$. This opens
two questions we study together: whether the richer \textsc{deprel} prior helps,
and, once it is combined with the terminal and non-terminal priors, \emph{how}
all of these vectors should be fused with the token and positional embeddings,
by addition, concatenation, a learned interpolation weight, or injection deeper
into the network.

\begin{figure*}[t]
    \centering
    \includegraphics[width=0.95\textwidth]{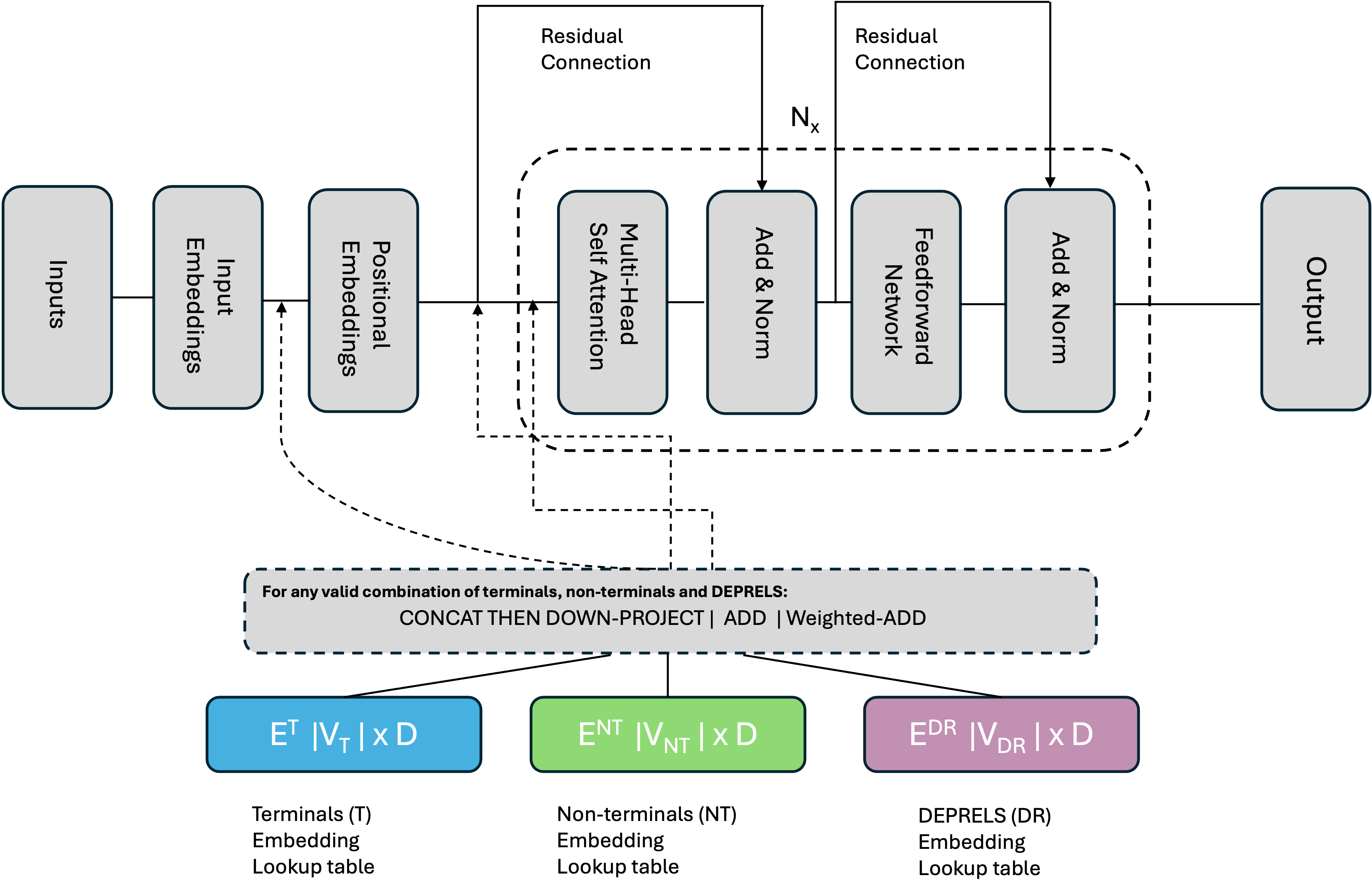}
\caption{\textbf{Injection points for syntactic priors in a Transformer encoder.}
    We study where and how to inject hexatag-derived priors (terminal $\mathbf{t}_i$, non-terminal $\mathbf{n}_i$, and optionally dependency-relation labels $\mathbf{deprel}_i$) into RoBERTa-base. Three architectural locations are considered: (i) at the \emph{input embedding}, alongside the token, segment, and positional embeddings; (ii) at the \emph{residual (skip) connection} entering the first self-attention sublayer; and (iii) implicitly in the attention computation, by injecting at the input and propagating through $\mathbf{W}_Q, \mathbf{W}_K$.}
    \label{fig:ablation}
\end{figure*}

We explore both questions as a small neural-architecture search over injection
strategies. To keep it tractable we run the search on the simplest positional
scheme, absolute positional embeddings, using \texttt{RoBERTa-base} as the
encoder (the case illustrated in Figure~\ref{fig:ablation}). The strategies we
explore are collected in Table~\ref{tab:injection_strategies} and split into two
studies at different pretraining scales.

\begin{table*}[t]
\centering
\small
\setlength{\tabcolsep}{6pt}
\renewcommand{\arraystretch}{1.2}
\begin{tabularx}{\textwidth}{@{}l l >{\raggedright\arraybackslash}X@{}}
\toprule
\textbf{Strategy} & \textbf{Injected input representation $\mathbf{x}'_i$} & \textbf{Description} \\
\midrule
\multicolumn{3}{@{}l}{\textit{Fusion mechanism (terminal + non-terminal tags)}} \\
\textbf{ADD\_T\_NT} & $\mathbf{x}'_i = \mathbf{x}_i + \mathbf{t}_i + \mathbf{n}_i$ & Add both priors to the input embedding (default; no extra parameters). \\
\textbf{CONCAT\_T\_NT} & $\mathbf{x}'_i = \mathbf{W}\,[\,\mathbf{x}_i;\, \mathbf{t}_i;\, \mathbf{n}_i\,]$ & Concatenate with the input embedding, then down-project to $D$. \\
\textbf{ADD\_T\_NT (Residual Connection)} & $\mathbf{h}^{(1)}_i = \mathrm{LN}\bigl(\mathbf{x}_i + \mathbf{t}_i + \mathbf{n}_i + \mathrm{Attn}(\mathbf{x})_i\bigr)$ & Add the priors to the skip connection of the first attention sublayer. 
\\
\textbf{ADD\_T\_NT ($\alpha$)} & $\mathbf{x}'_i = \alpha\,\mathbf{x}_i + (1-\alpha)(\mathbf{t}_i + \mathbf{n}_i)$ & Linearly interpolate the input embedding with the (terminal + non-terminal) prior; learned $\alpha$ initialized in $\{0.1,0.5,0.9\}$. \\
\midrule
\multicolumn{3}{@{}l}{\textit{Adding dependency relations ($\mathbf{deprel}_i$, 40 labels)}} \\
\textbf{ADD\_T\_NT\_DR} & $\mathbf{x}'_i = \mathbf{x}_i + \mathbf{t}_i + \mathbf{n}_i + \mathbf{deprel}_i$ & Add all three priors to the input embedding. \\
\textbf{CONCAT\_POS\_NT\_T\_DR} & $\mathbf{x}'_i = \mathbf{W}\,[\,\mathbf{x}_i;\, \mathbf{n}_i;\, \mathbf{t}_i;\, \mathbf{deprel}_i\,]$ & Concatenate all priors with the input embedding, then down-project. \\
\textbf{ADD\_DNT} & $\mathbf{x}'_i = \mathbf{x}_i + \mathbf{deprel}_i + \mathbf{n}_i$ & Swap the terminal tag for the \textsc{deprel}. \\
\textbf{ADD\_DNT ($\alpha$)} & $\mathbf{x}'_i = \alpha\,\mathbf{x}_i + (1-\alpha)(\mathbf{deprel}_i + \mathbf{n}_i)$ & Interpolated counterpart of \textbf{ADD\_DNT}; learned $\alpha$ initialized in $\{0.1,0.5,0.9\}$. \\
\textbf{CONCAT\_POS\_DNT} & $\mathbf{x}'_i = \mathbf{W}\,[\,\mathbf{x}_i;\, \mathbf{deprel}_i;\, \mathbf{n}_i\,]$ & Concatenate (\textsc{deprel} replacing the terminal), then down-project. \\
\bottomrule
\end{tabularx}
\caption{Injection strategies we explore for absolute positional embeddings on
\texttt{RoBERTa-base}. $\mathbf{x}_i = \mathbf{e}_i + \mathbf{p}_i$ is the
standard input embedding (token $\mathbf{e}_i$ plus positional $\mathbf{p}_i$),
$\mathbf{x}'_i$ the representation actually fed to the encoder, and $\mathbf{W}$
a learned projection back to the model dimension $D$. $\mathbf{t}_i$/$\mathbf{n}_i$
are the terminal/non-terminal hexatag priors and $\mathbf{deprel}_i$ the
dependency-relation prior. For the residual variant we write the first
attention sublayer explicitly, $\mathbf{h}^{(1)} = \mathrm{LN}(\mathbf{x} +
\mathrm{Attn}(\mathbf{x}))$: the priors are added on the skip branch, so unlike
\textbf{ADD\_T\_NT} they do not enter the self-attention computation itself.
The top block fuses only the terminal and
non-terminal tags; the bottom block additionally brings in \textsc{deprels}.}
\label{tab:injection_strategies}
\end{table*}

\paragraph{Fusion mechanism (1M steps).}
We first fix the priors to $\mathbf{t}_i$ and $\mathbf{n}_i$ and vary only
\emph{how} they enter the model, pretraining \texttt{RoBERTa-base} for 1M steps
(top block of Table~\ref{tab:injection_strategies},
Table~\ref{tab:other_combinations}). The simplest option wins:
\textbf{ADD\_T\_NT}, which just adds the two priors to the input embedding with
no extra parameters, attains the best macro-average. \textbf{CONCAT\_T\_NT} is
the weakest (with large drops on QNLI and CoLA), and injection into the model's residual (skip) connection
(\textbf{ADD\_T\_NT (Residual Connection)}) yields only modest gains. Replacing the plain
addition with a learned linear interpolation between the input embedding and the
prior, \textbf{ADD\_T\_NT ($\alpha$)}: $\mathbf{x}'_i = \alpha\,\mathbf{x}_i +
(1-\alpha)(\mathbf{t}_i+\mathbf{n}_i)$, never improves over
\textbf{ADD\_T\_NT} at any $\alpha$ initialization
(Table~\ref{tab:alpha_combinations}). Adding capacity or depth to the fusion,
therefore, does not help; direct addition at the embedding layer is both the
cheapest and the strongest.

\paragraph{Adding dependency relations (500k steps).}
We then enrich the prior itself: we embed each token's \textsc{deprel} label and
combine it with the terminal and non-terminal priors through the bottom-block
strategies of Table~\ref{tab:injection_strategies}. Because this study is more
expensive, every variant, including a re-run baseline and \textbf{ADD\_T\_NT},
is pretrained for 500k steps for an apples-to-apples comparison
(Figures~\ref{fig:glue_avg_ablation}, \ref{fig:deprel_per_dataset}). Every
prior-injection variant improves over the no-prior baseline on macro-average,
yet \textbf{ADD\_T\_NT} still attains the highest mean. Layering \textsc{deprels}
on top of $\mathbf{t}_i+\mathbf{n}_i$ (\textbf{ADD\_T\_NT\_DR},
\textbf{CONCAT\_POS\_NT\_T\_DR}) does not close the gap, and swapping the
terminal tag for the \textsc{deprel} is no better, whether added directly
(\textbf{ADD\_DNT}), concatenated (\textbf{CONCAT\_POS\_DNT}), or interpolated
through a learned weight, \textbf{ADD\_DNT ($\alpha$)}:
$\mathbf{x}'_i = \alpha\,\mathbf{x}_i + (1-\alpha)(\mathbf{deprel}_i+\mathbf{n}_i)$.
Richer syntactic signal, at least in this form, does not translate into stronger
downstream performance.

\paragraph{Summary and scope.}
Taken together, these experiments identify \textbf{ADD\_T\_NT}, direct addition
of the terminal and non-terminal priors at the input embedding, as the strongest
injection strategy for absolute positional embeddings on the \texttt{RoBERTa}
encoder. Two considerations bound the generality of this finding. First, the
per-dataset results (Figure~\ref{fig:deprel_per_dataset}) show that no single
variant is optimal on every task; the relative ordering of strategies varies
across GLUE tasks, and \textbf{ADD\_T\_NT} is strongest in the macro-average
rather than uniformly. Second, our search covers only the absolute-PE encoder;
the optimal injection strategy for relative (DeBERTa-v3, Transformer-XL) and
rotary (ModernBERT) positional schemes, and under larger pretraining budgets,
remains open. We therefore adopt \textbf{ADD\_T\_NT} as a simple and robust
default, and leave a systematic study of injection strategies across positional
schemes to future work.

